\documentclass[10pt]{article} 

\usepackage[preprint]{rlj} 

\usepackage{amssymb}            
\usepackage{mathtools}          
\usepackage{mathrsfs}           
\usepackage{graphicx}           
\usepackage{subcaption}         
\usepackage[space]{grffile}     
\usepackage{url}                
\usepackage{lipsum}             
\usepackage{algorithm}
\usepackage{algpseudocode}
\usepackage{bbm}
\usepackage{booktabs}
\title{Intrinsic Vicarious Conditioning for Deep Reinforcement Learning}

\setrunningtitle{Vicarious Conditioning for DRL}

\author{Rodney A Sanchez\textsuperscript{1}, Ferat Sahin\textsuperscript{1}, Alex Ororbia\textsuperscript{2}, Jamison Heard\textsuperscript{1} }

\emails{{ras8047,feseee,agovcs,jrheee}@rit.edu}

\affiliations{
$^{1}$\textbf{Department of Electrical and Microelectronic Engineering, Rochester Institute of Technology}\\
$^{2}$\textbf{Department of Computer Science, Rochester Institute of Technology}\\
\par 

}

\contribution{
 We provide a general formulation for vicarious conditioning within the framework of RL, grounded in Bandura's four-step model of observational learning. 
    }
    {
    None
    }

\contribution{
We design and introduce a Siamese memory-augmented neural network architecture whose shared encoder ensures that behaviorally similar sequences produce comparable embeddings across demonstrator and agent, enabling vicarious conditioning under low-shot learning conditions.
    }
    {Biological vicarious conditioning is an online learning method; our formulation accounts for this, but our memory implementation is an offline method. 
    }

\contribution{
 We craft mechanisms to engender both negative and positive vicarious conditioning: negative conditioning for avoidance, and positive conditioning for approach.
    }
    {None

    }


\summary{Advancements in reinforcement learning have produced a variety of complex and useful intrinsic driving forces; crucially, these drivers operate under a direct conditioning paradigm. This form of conditioning limits our agents’ capacity by restricting how they learn from the environment as well as from others. Off-policy or learn-by-example methods can learn from demonstrators' representations, but they require access to the demonstrating agent’s policies or their reward functions. Our work overcomes this direct sampling limitation by introducing vicarious conditioning as an intrinsic reward mechanism. We draw from psychological and biological literature to provide a foundation for vicarious conditioning and use memory-based methods to implement its four steps: 
attention, retention, reproduction, and reinforcement. 
Crucially, our vicarious conditioning paradigms support low-shot learning and do not require the demonstrator agent's policy nor its reward functions. We evaluate our approach in the MiniWorld Sidewalk environment, one of the few public environments that features a non-descriptive terminal condition (no reward provided upon agent death), and extend it to Box2D’s CarRacing environment through a custom partially observable Markov decision process wrapper. Our results across both environments demonstrate that vicarious conditioning enables longer episode lengths by discouraging the agent from non-descriptive terminal conditions. Collectively, our experiments show that vicarious conditioning provides a mechanism for integrating observed demonstrations into the policy in order to guide an agent’s behavior, either through avoidance or enticement, using only a few demonstrated trajectories. Overall, this work emulates a cognitively-plausible learning paradigm better suited to problems such as single-life learning or continual learning.
}

\begin{document}
\maketitle  

\begin{abstract}
Advancements in reinforcement learning have produced a variety of complex and useful intrinsic driving forces; crucially, these drivers operate under a direct conditioning paradigm. This form of conditioning limits our agents’ capacity by restricting how they learn from the environment as well as from others. Off-policy or learn-by-example methods can learn from demonstrators' representations, but they require access to the demonstrating agent’s policies or their reward functions. Our work overcomes this direct sampling limitation by introducing vicarious conditioning as an intrinsic reward mechanism. We draw from psychological and biological literature to provide a foundation for vicarious conditioning and use memory-based methods to implement its four steps: 
attention, retention, reproduction, and reinforcement. 
Crucially, our vicarious conditioning paradigms support low-shot learning and do not require the demonstrator agent's policy nor its reward functions. We evaluate our approach in the MiniWorld Sidewalk environment, one of the few public environments that features a non-descriptive terminal condition (no reward provided upon agent death), and extend it to Box2D’s CarRacing environment. Our results across both environments demonstrate that vicarious conditioning enables longer episode lengths by discouraging the agent from non-descriptive terminal conditions and guiding the agent toward desirable states.
Overall, this work emulates a cognitively-plausible learning paradigm better suited to problems such as single-life learning or continual learning.

\end{abstract}
\section{Introduction} 
\label{sec:intro}

Reinforcement learning (RL) has seen significant advances in reward shaping methods that aim to facilitate learning in complex real-world environments. Collectively, these methods leverage expert behavior either through a demonstrator's policy or its reward function in order to promote better alignment between the training agent and the demonstrator~\citep{pmlr-v205-guo23a,pmlr-v162-liu22u,10342230}. 
However, these approaches share a common limitation: they require access to the expert's internal representations, action-level trajectories, or knowledge of the demonstrator's reward function. This requirement becomes problematic in environments where the extrinsic reward is sparse and where catastrophic terminal conditions carry no reward signal, e.g., death. In these contexts, the agent must repeatedly experience failure before its policy can adapt in such settings.

Biological agents face analogous constraints -- for instance, partial observability, sparse feedback, and costly failure -- yet they routinely learn to avoid danger and pursue desirable behaviors without direct experience. In essence, these agents accomplish this through \textit{vicarious conditioning}, a process by which an agent learns from observing another agent's behavior and its communicated outcome, all without access to the demonstrator's reward function or policy~\citep{Askew2008TheOn}. Vicarious conditioning can be decomposed into four distinct steps: attention, retention, reproduction, and reinforcement~\citep{Szczepanik2020-fu}. 
Attention and retention govern which features the agent encodes and stores from the observed behavior. Reproduction enables the agent to recognize when its own experience resembles a stored behavior and results in it generating an intrinsic reward signal accordingly. Reinforcement allows the environment to confirm or override the vicariously learned values, yielding a policy that reflects the demonstrator's guidance but remains optimal for the agent. 

Our work emulates the above biological framework by introducing vicarious conditioning as an intrinsic reward mechanism for RL. Our method, as a result, only requires a few demonstrated trajectories along with the demonstrator's communicated evaluation of the outcome; this does not require the demonstrator's policy, reward function, or action labels. We evaluate our approach on the MiniWorld Sidewalk environment, one of the few public environments with a non-descriptive terminal condition and then extend it to Box2D's CarRacing environment via a partially observable Markov decision process (POMDP) wrapper. In summary, the key contributions of this work are: 
\begin{itemize}[noitemsep,nolistsep]
    \item We provide a general formulation for vicarious conditioning within the framework of RL, grounded in Bandura's four-step model of observational learning. 
    \item We design and introduce a Siamese memory-augmented neural network architecture whose shared encoder ensures that behaviorally similar sequences produce comparable embeddings across demonstrator and agent, enabling vicarious conditioning under low-shot learning conditions.
    \item We craft mechanisms to engender both negative and positive vicarious conditioning: negative conditioning for avoidance, and positive conditioning for approach. 
\end{itemize}

As a result, our method takes a step toward a single-life RL \citep{chen2022you} paradigm, where agents cannot afford to learn about catastrophic failure through direct experience. By enabling agents to learn from a few observed demonstrations, as opposed to self-exploration, this works shows that vicarious conditioning offers a cognitively-inspired alternative better suited to real-world environments that are vast, partially observable, and dangerous.

\section{Related Work}
\label{sec:related_work}

\subsection{Reward Shaping}
\label{subsec:reward_shaping}

Reward shaping methods augment the agent's learning signal to address partial observability, sparse / delayed rewards, and long-horizons. Custom reward functions address complex real-world scenarios by encoding expert knowledge directly into the reward signal, seeking to balance multiple goals to produce optimal incentives for the policy~\citep{8968478,9294338,8950131}. These schemes produce dense rewards for long-horizon tasks (e.g., driving~\citep{wang2023efficient}) or formulate rewards around salient state representations~\citep{Xing2021}. 
All of these approaches require extensive expert knowledge to design and are tailored to specific environments.

\subsection{Learning from Demonstrations and Observations}
\label{subsec:learning_from_demos}
Learning from demonstration (LfD) methods seek to train agents using expert-provided, state-action trajectories. Behavioral cloning (BC) treats the problem as supervised learning over state-action pairs~\citep{pomerleau1991efficient}, though it suffers from compounding errors ~\citep{ross2011reduction}. Inverse RL recovers the expert's reward function from demonstrations~\citep{abbeel2004apprenticeship,ziebart2008maximum} whereas adversarial formulations, such as GAIL~\citep{NIPS2016_cc7e2b87} and AIRL~\citep{fu2018learning}, unify imitation and reward recovery. These methods have been extended to handle policy alignment~\citep{pmlr-v162-kang22a,pmlr-v205-guo23a,pmlr-v162-xu22l}, representation processing~\citep{pmlr-v162-hansen-estruch22a,pmlr-v139-yarats21a,10342230}, and learning from suboptimal or mismatched demonstrations~\citep{pmlr-v162-liu22u,9340915,Deep_Q_from_demonstration,9285219}. Across all of these methods, one common requirement persists: access to the expert's actions, large demonstration datasets, or recovery of the expert's reward function.

Learning from observation (LfO) relaxes the action requirement by training agents from observation-only trajectories~\citep{torabi2019recent}, inferring missing actions through inverse dynamics models~\citep{torabi2018behavioral}. Other methods, in contrast, learn reward functions or domain-translation mappings from observation sequences~\citep{liu2018imitation,stadie2017third}. LfO methods typically aim to recover or approximate the demonstrator's policy, often requiring many trajectories.

Vicarious conditioning shares LfO's observation-only constraint yet differs in both objective and mechanism. Notably, our method does not attempt to recover the demonstrator's policy or reward function. Instead, the agent stores a few demonstrated trajectories, alongside a communicated outcome value, and generates an intrinsic reward based on the similarity between its current behavior and stored demonstrations. The agent's policy is shaped by this intrinsic signal but optimized under its own reward function. This distinction is crucial: vicarious conditioning produces agents that learn \textit{about} the demonstrator's experience rather than learning \textit{to replicate} it.

\subsection{Psychological/Biological Basis}
\label{subsec:psych}
Vicarious conditioning can be decomposed into fear and anxiety (negative reward) and anticipation (positive reward), with four distinct steps~\citep{Askew2008TheOn} : 
\textit{attention}, \textit{retention}, \textit{reproduction}, and \textit{reinforcement}. 
Fear acquisition requires sensory processing and storage in memory (\textit{attention} and \textit{retention})~\citep{Martin2009-uk,Fear_Synapse_Growth,DESMEDT2015290,LANZILOTTO2025106006}. The amygdala then performs value estimation (\textit{reproduction, reinforcement)}), either for positive (e.g., addiction literature~\citep{Koob2016-rt,Johnson2025-kb,LUO2013159}) or negative values as seen in fear and anxiety literature~\citep{Laufer2016-qu,GHAEMIKERAHRODI2020101588,Martin2009-uk}. 
Stimulus avoidance emerges before amygdala development, while behavior reproduction does not~\citep{Sullivan2020-ho}. Vicarious conditioning is further modulated by inhibition, where the agent can alter the value of a previously learned behavior~\citep{JOVANOVIC20051559,HERMANS2006361,exposure_therapy_fear,fear_memory_psycology_treatment}.
Vicarious conditioning further shifts from a parent-to-child format to a peer-to-peer one as the agent enters adolescence~\citep{Marin2020VicariousDyads,Hostinar2013-hh}. Parent-to-child conditioning demonstrates that children undergo vicarious conditioning from the parent even without directly experiencing negative stimuli (e.g., shock), an effect not observed with strangers~\citep{Marin2020VicariousDyads}. This is referred to, as in \citet{Hostinar2013-hh}, as an inherent trust in parental conditioning as social buffering, where the child modulates their response to match the parent's (an effect that diminishes post-adolescence). Mechanistically, the biological / cognitive literature demonstrates that vicarious conditioning requires memory to distinguish behaviors and stimuli (hippocampus), value estimation over outcomes (amygdala), and an inhibition mechanism that gates the response (prefrontal cortex). 

\subsection{Memory Methods}
\label{subsec:memory_anns}
Memory-augmented neural networks (MANNs) have proven to be particularly useful for leveraging external memory for low-shot and few-shot classification problems~\citep{Hopfield1982-cp, Santoro2016Meta-LearningNetworks}. This architecture uses neural Turing machines to access memory through content-based addressing~\citep{graves2014neural}; the MANN model was further expanded to work with convolutional network structures to acquire separable representations capable of distinguishing between different categories~\citep{Mao2022}. Modern variants of memory methods use transformers, taking advantage of their self-attention-driven, key-value querying mechanism to further encapsulate representations. However, these transformer-based approaches require large datasets to train well and struggle to perform low-shot learning~\citep{10.1007/978-3-031-19815-1_37}. 

\section{Problem Formulation}
\label{sec:problem_form}

We consider a partially observable Markov decision process (POMDP) defined by the tuple $\mathcal{M} = (\mathcal{S}, \mathcal{A}, \mathcal{O}, P, O, r, \gamma)$, where $\mathcal{S}$ is the state space, $\mathcal{A}$ is the action space, $\mathcal{O}$ is the observation space, $P(s' \mid s, a)$ is the transition function, $O: \mathcal{S} \rightarrow \mathcal{O}$ is the emission function, $r: \mathcal{S} \times \mathcal{A} \rightarrow \mathbb{R}$ is the reward function, and $\gamma \in [0,1)$ is the discount factor. 
At each timestep $t$, the agent observes $o_t=O(s_t)$, selects $a_t \sim \pi(o_t)$, and the environment transitions to $s_{t+1} \sim P(\cdot \mid s_t, a_t)$. The agent's goal is to learn a policy ($\pi: \mathcal{O} \rightarrow \Delta(\mathcal{A})$) that maximizes the expected return $\mathbb{E}_\pi\left[\sum_{t=0}^{\infty} \gamma^t r_t\right]$.

We focus on partially observable environments where the extrinsic (environmental) reward ($r^{\text{ext}}$) is sparse and where some terminal conditions are non-descriptive ($r^{\text{ext}} = 0$). These conditions are motivated from realistic settings where a sparse catastrophic event (e.g., death) may occur without any feedback. 
We address these conditions using \textit{vicarious conditioning (VC)} where an intrinsic reward ($r^{vic}$) permits agents to learn states from observing another agent's trajectory without direct access to that agent's reward function. As a result, the agent optimizes the combined reward function: 
\begin{equation}
\label{eq:composite_reward}
    \bar{r}(o_t) = r^{\text{ext}}(o_t) + \alpha \cdot r^{\text{vic}}(o_t),
\end{equation}
where $\alpha > 0$ controls the vicarious reward's influence. Unlike reward shaping, the vicarious reward is not derived from a potential function over states. This VC formulation biases the agent away from dangerous states and towards desirable state trajectories that the sparse environmental reward does not adequately describe (where another agent has been observed performing behaviors).

\textbf{Demonstrator Trajectory.} The agent observes a demonstrator trajectory $\tau = (o_1^d, o_2^d, \ldots, o_T^d, v_\tau)$, where $o_t^d \in \mathcal{O}$ are the demonstrator's observations and $v_\tau \in \mathbb{R}$ is a \textit{communicated value} that expresses the demonstrator's assessment of the trajectory's outcome. The sign of $v_\tau$ indicates valence (negative for avoidance, positive for approach) and its magnitude indicates intensity.
Crucially, the agent does not observe the demonstrator's reward function $r^d$ or policy $\pi^d$; it only observes the trajectory and communicated valence (negative or positive). In addition, the agent does not observe the demonstrator's internal representations. This framing aligns with biological vicarious conditioning, where a parent's reaction communicates both the valence and intensity of an outcome to the child.

\section{Vicarious Conditioning}
\label{sec:vicarious_conditioning}
Vicarious conditioning is decomposed into four distinct steps: 
\textit{attention}, \textit{retention}, \textit{reproduction}, and \textit{reinforcement}. We formalize each step within an RL framework, where we show the theoretical formulation and our implementation using a Siamese memory-augmented neural network. 
\subsection{Attention} 
\label{subsec:attention}

Attention governs which aspects of the demonstrator's behavior the agent can perceive / encode within two constraints: 
\textbf{(1)} partially observability, and 
\textbf{(2)} the agent must extract relevant features / representations from variable-length sequences. 
Under partially observability, the agent generates a belief state from observations (rather than direct states). Thus, the agent receives a sequence of observations ($\tau$) that constitutes a partial view of the demonstrator's behavior.

A \textit{controller encoder} 
 maps an observation sequence to a fixed-dimensional embedding: 
\begin{equation}
    \mathbf{e}_\tau = \text{CE}(o_1^d, o_2^d, \ldots, o_T^d),
    \label{eq:controller_encoding}
\end{equation}
where $\mathbf{e}_\tau \in \mathbb{R}^d$ is the encoding of the observed behavior. This encoder is applied to the agent's own observation window of length $l$, producing $\mathbf{e}_t = \text{CE}(o_{t-l+1}, \ldots, o_t)$. Critically, the same encoder is used for the demonstrator and agent to ensure that similar behaviors produce similar embeddings. 
Note that the demonstrator does not have access to this encoder; hence, only the agent uses it.

\textbf{Implementation:} We implement $\text{CE}$ as a Siamese long-short-term memory (LSTM) \cite{hochreiter1997long,varior2016siamese} recurrent network where the standard linear layers are replaced with a shared network $\mathcal{G}$ that jointly processes the current observation, prior hidden state, and reward context. This architecture enables the controller to process variable-length sequences while maintaining a consistent encoding across behaviors.

\subsection{Retention}
\label{subsec:retention}

Retention describes the process by which an agent stores observed behaviors ($\tau$) alongside their communicated value ($v_\tau$). Given a demonstrator trajectory ($\tau$), the controller encoder (Eq. \ref{eq:controller_encoding}) produces a fixed dimensional embedding; 
this stores the behavior representation. Retention then involves 
writing (retaining) the encoded representations to external memory.

\textbf{Vicarious Value Attribution.} For each time step along 
$\tau$, we define the vicarious value via backward discounting from the outcome utility, modulated by the trust coefficient ($k \in [0,1]$) as:
\begin{equation}
    V^{\text{vic}}(o_t^d) = k \cdot \gamma^{T-t} \cdot v_\tau.
    \label{eq:vicarious_value}
\end{equation} 
Representations closer to the outcome carry larger magnitude, while $k$ governs how much the agent trusts the demonstrator / trajectory. The above vicarious value formulation does not require knowledge of the demonstrator's reward function; 
$\gamma^{T-t}$ emerges from the representation geometry and is not computed explicitly. In this work, we assume $k = 1$, consistent with a parent-child trust paradigm (Section~\ref{subsec:psych}). Exploration of $k < 1$ for peer-to-peer pairings is left to future work. 

\textbf{Memory Writing.} The embedding ($e_\tau$) and its associated vicarious value are written to external memory $\mathbf{M}_t \in \mathbb{R}^{N \times d}$ with $N$ memory slots via content-based addressing using the least recently used access value via: $\mathbf{M}_t(i) \leftarrow \mathbf{M}_{t-1}(i) + w_t^w(i) \cdot e_\tau, \quad \forall i,$ 

where $w_t^w$ is the write vector and each slot is associated with the communicated vicarious value $v_{\tau}$. While only $e_\tau$ is written to the memory matrix, $v_\tau$ serves as the supervisory signal during training and determines the reward valence.

\textbf{Implementation:} We utilize a MANN as external memory, due to being an end-to-end (CE, memory addressing, and output) model. The MANN is trained with a cross-entropy loss over the vicarious values, where $v_\tau \rightarrow \omega \in \Omega$  discretizes the VC values. 
Concretely, we use $|\Omega| = 2$ classes: a negative class ($\omega^-$, $v_\tau < 0$) for avoidance demonstrations, and a positive class ($\omega^+$, $v_\tau > 0$) for approach demonstrations. 
In single-valence environments, only the corresponding class is populated in memory. The MANN's parameters are frozen ensuring the vicarious reward signal remains stationary when training the agent.

\subsection{Reproduction}
\label{subsec:reproduction}

Reproduction captures the agent's ability to recognize when its current trajectory resembles a stored behavior to generate an appropriate intrinsic reward signal.

\textbf{Generalization Kernel.} We define a generalization kernel $K: \mathbb{R}^d \times \mathbb{R}^d \rightarrow [0, 1]$ that operates in a shared embedding space provided by the CE (Sec. \ref{subsec:attention}): 
\begin{equation}
    K(\mathbf{e}_t, \mathbf{e}_\tau) = \frac{\exp\left(\cos(\mathbf{e}_t, \mathbf{e}_\tau)\right)}{\sum_j \exp\left(\cos(\mathbf{e}_t, \mathbf{e}_j)\right)},
    \label{eq:generalization_kernel}
\end{equation}
where $\cos(\cdot, \cdot)$ denotes the cosine similarity and the normalization is applied over all stored memory entries ($j$). 
Since each embedding is produced by the same encoder, behaviorally similar sequences will produce high similarity (kernel) scores.

\textbf{Memory Read and Classification.} The read weights ($w_t^r(i) = K(\mathbf{e}_t, \mathbf{M}_t(i))$) yield a retrieved memory $\mathbf{m}_t = \sum_i w_t^r(i) \cdot \mathbf{M}_t(i)$. The MANN then classifies the agent's current behavior as follows:

\begin{equation}
    \mathbf{p}_t = \text{softmax}\left(\mathbf{W} \begin{bmatrix} \mathbf{e}_t \\ \mathbf{m}_t \end{bmatrix} + \mathbf{b}\right) \in \mathbb{R}^{|\Omega|},
    \label{eq:class_probs}
\end{equation}
where, $\mathbf{W}$ and $\mathbf{b}$ are learned parameters and $p_t$ is the confidence that a behavior resembles a stored demonstration.

\textbf{Vicarious Intrinsic Reward.} The vicarious intrinsic reward is aggregated from all $p_t$ gated via an inhibition scheme that helps prevent activation(s) from low-confidence matches: 
\begin{equation}
    r^{\text{vic}}(o_t) = \sum_{\omega \in \Omega} \mathbf{1}\left[p_{t,\omega} > \theta_{\text{thr}}\right] \cdot v_{\tau,\omega} \cdot p_{t,\omega},
    \label{eq:intrinsic_reward}
\end{equation}
where $\Omega$ is the set of behavior classes demonstrated 
and $\theta_{\text{thr}} \in [0, 1]$ is the inhibition threshold. Each behavior class contributes independently: 
if the agent's state strongly resembles a negative class trajectory, then the agent receives a high negative reward proportional to $v_{\tau,\omega}$. Conversely, if there is not a strong resemblance, the agent does not receive any reward signal (inhibited). 

\textbf{Implementation.} The output layer parameters $\mathbf{W}$ and $\mathbf{b}$ (Eq.~\ref{eq:class_probs}) are trained jointly with the MANN during the supervised phase described in Section~\ref{subsec:retention}. At inference time, each forward pass through the frozen MANN produces both the retrieved memory $\mathbf{m}_t$ and the classification probabilities $\mathbf{p}_t$ within a single step, with no additional learned components required during the RL agent's training.

\subsection{Reinforcement}
\label{subsec:reinforcement}

The intrinsic reward $r^{\text{vic}}(o_t)$ produced by the reproduction step (Eq.~\ref{eq:intrinsic_reward}) is combined with the extrinsic reward via the composite reward via: 
$\bar{r}(o_t) = r^{\text{ext}}(o_t) + \alpha \cdot r^{\text{vic}}(o_t)$ (Eq.~\ref{eq:composite_reward}). 
Under this composite reward, the agent learns an observation-conditioned value function \citep{sutton1998introduction}:
\begin{equation}
    V(o_t) \approx \max_a \left[ r^{\text{ext}}(o_t, a) + \alpha \cdot r^{\text{vic}}(o_t) + \gamma\, \mathbb{E}_{o_{t+1}} \left[ V(o_{t+1}) \right] \right].
    \label{eq:bellman_composite}
\end{equation} 
This formulation reinforces the vicarious conditioning when the extrinsic reward confirms the vicarious signal: 
negative valence leads to avoidance behavior, whereas a positive valence further incentivizes the extrinsic reward. These conditions are only satisfied when $|r^{\text{ext}}_{\max}| > |\alpha \cdot r^{\text{vic}}_{\max}|$; this ensures that standard policy optimization favors the extrinsic signal. If a contradiction occurs, then the extrinsic reward eventually dominates. Pseudocode is provided in supplementary materials.

\section{Experiments} 
\label{sec:experiments}

We evaluate vicarious conditioning in two environments that test complementary arms of our framework. The MiniWorld Sidewalk environment tests negative conditioning (avoidance learning) under a non-descriptive terminal condition; Box2D's CarRacing environment tests positive conditioning (approach learning) with a tile-based extrinsic reward. For both environments, we compare against base PPO (extrinsic reward only) and the stimuli-based baseline of \citet{sanchez2024fear}, which applies a continuous intrinsic penalty without threshold gating. 
All results are computed over the final $200$ training episodes across $5$ runs. We report $95$\% confidence intervals (CIs) via the $t$-distribution throughout; statistical comparisons use Welch's $t$-test with Cohen's $d$ for effect size. We note that effect sizes are computed from $5$ runs per condition; reported Cohen's $d$ values reflect within-condition variance across those runs and should be interpreted as measures of separation relative to observed variability rather than universal practical effect magnitudes. 
Hyperparameters for all conditions are provided in the supplementary materials.

\subsection{Sidewalk Environment}
\label{subsec:sidewalk_env}

Table~\ref{tab:sidewalk_result_table} reports performance over the final $200$ training episodes across $5$ runs; training curves for a priori chosen representative thresholds appear in Figure~\ref{fig:Intrinsic_Only} and in supplementary materials. The dataset size (number of demonstrations) was $26$ with a trajectory step size of $3$. 
Base PPO achieves $116.88$ ($95$\% CI $[111.71, 122.05]$) steps with zero intrinsic and extrinsic reward. The low variance ($\pm 4.16$) reflects the environment geometry: the sidewalk is wide enough such that random exploration sustains moderate survival, but the non-descriptive terminal condition provides no gradient for learning avoidance. 
The stimuli baseline~\citep{sanchez2024fear} performs significantly worse ($t(4.2) = 5.59$, $p = .004$, $d = 3.54$), achieving only $44.30$ ($95$\% CI $[8.65, 79.95]$) steps. This un-gated intrinsic penalty approximates a living cost: every time step incurs punishment, so under a non-descriptive terminal condition the agent learns to minimize episode length in order to reduce cumulative penalty. This shows that naive intrinsic fear without threshold gating is actively harmful.

\begin{table}[htb]
\centering
\caption{Sidewalk environment results over the final $200$ training episodes across $5$ runs (mean $\pm$ SD). VC: vicarious conditioning.}
\label{tab:sidewalk_result_table}
\resizebox{0.90\columnwidth}{!}{%
\begin{tabular}{lccc}
\hline
Method & Episode Length & Intrinsic Reward & Extrinsic Reward \\ \hline
\multicolumn{4}{c}{\textbf{Baselines}} \\ \hline
Base PPO & $116.88 \pm 4.16$ & $0$ & $0$ \\
PPO + Stimuli~\citep{sanchez2024fear} & $44.30 \pm 28.71$ & $-21.81 \pm 15.84$ & $0.00 \pm 0.00$ \\ \hline
\multicolumn{4}{c}{\textbf{Intrinsic Reward Only}} \\ \hline
PPO + VC ($\theta_{\text{thr}} = 0.25$) & $130.59 \pm 13.53$ & $-2.09 \pm 2.20$ & $0.00 \pm 0.00$ \\
PPO + VC ($\theta_{\text{thr}} = 0.60$) & $\mathbf{139.96 \pm 11.17}$ & $\mathbf{-1.73 \pm 1.72}$ & $0.00 \pm 0.00$ \\
PPO + VC ($\theta_{\text{thr}} = 0.95$) & $100.23 \pm 50.26$ & $-16.49 \pm 17.46$ & $0.00 \pm 0.00$ \\ \hline
\multicolumn{4}{c}{\textbf{Composite Reward (Extrinsic + Intrinsic)}} \\ \hline
PPO + VC ($\theta_{\text{thr}} = 0.25$) & $126.38 \pm 14.86$ & $-5.71 \pm 9.58$ & $0.00 \pm 0.01$ \\
PPO + VC ($\theta_{\text{thr}} = 0.60$) & $\mathbf{142.16 \pm 7.72}$ & $\mathbf{-1.43 \pm 1.52}$ & $0.00 \pm 0.00$ \\
PPO + VC ($\theta_{\text{thr}} = 0.95$) & $133.80 \pm 12.46$ & $-1.84 \pm 2.95$ & $0.00 \pm 0.00$ \\ \hline
\end{tabular}%
}
\end{table}

Observe that VC at $\theta_{\text{thr}} = 0.60$ produces the strongest results. The composite reward achieves $142.16$ $[132.57, 151.75]$ steps, significantly outperforming both the base PPO ($t(6.1) = 6.45$, $p < .001$, $d = 4.08$) and the stimuli baseline ($t(4.6) = 7.36$, $p = .001$, $d = 4.66$). The intrinsic-only condition achieves comparable performance at $139.96$ $[126.09, 153.83]$ ($t(5.1) = 4.33$, $p = .007$, $d = 2.74$ vs.\ base PPO). The mid-range threshold facilitates accurate but not overly stringent classification of the demonstrated negative behavior: this discourages the agent from exploring states that approximate the demonstrator's trajectory and this avoidance incentivizes the agent to uncover safer paths that prolong survival. 
As shown in Figure~\ref{fig:Intrinsic_Only}a, the $\theta_{\text{thr}} = 0.60$ condition maintains episode length consistently above the base PPO reference throughout training with tight variance bands. 

\begin{figure}[!tb] 

   \centering
    \begin{subfigure}[t]{0.32\textwidth}
        \centering
        \includegraphics[width=0.9\textwidth, trim={0 0 0 1.2cm}, clip]{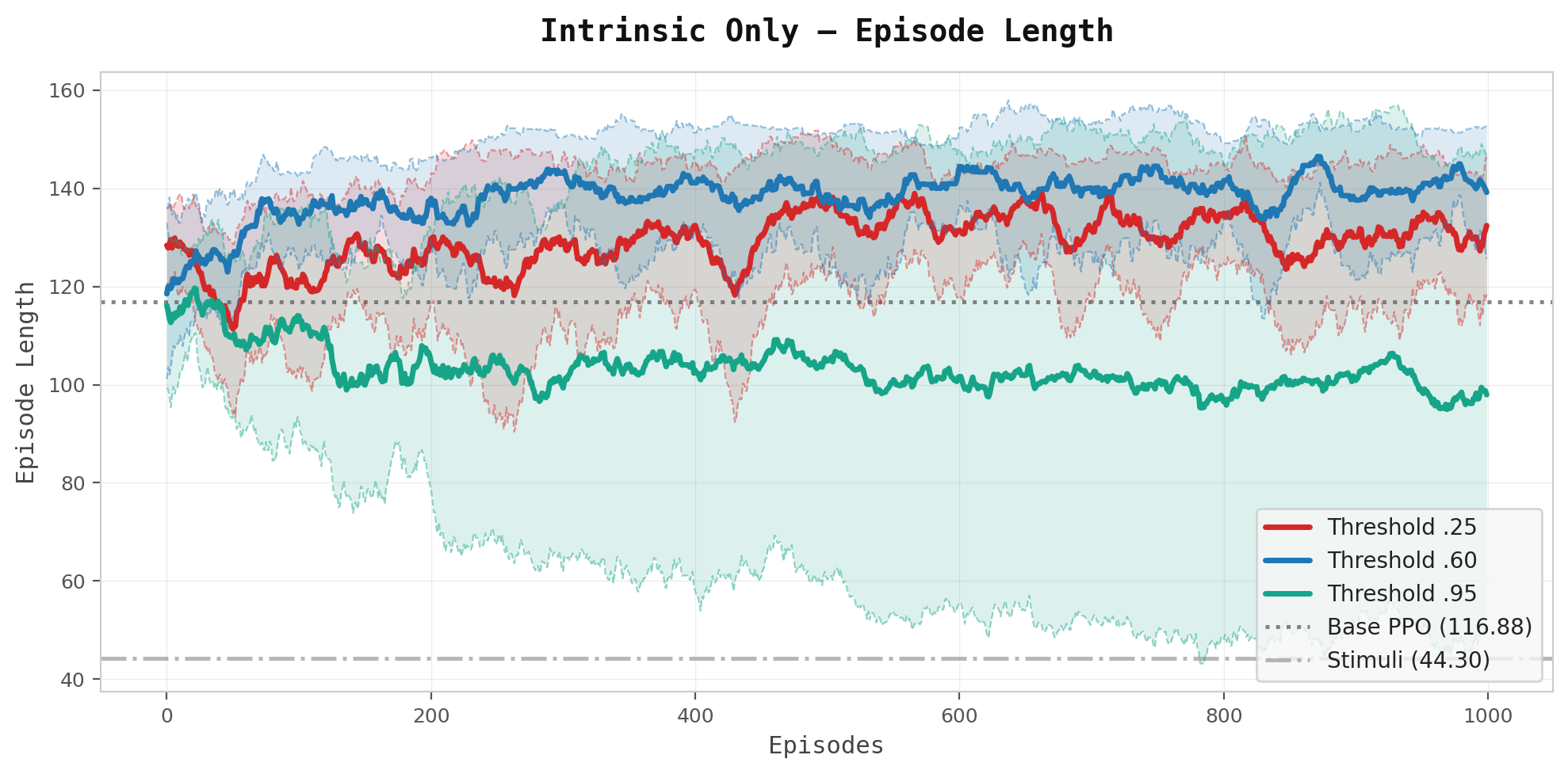}
        \caption{Episode Length}
        \label{fig:INT_LENGHT}
    \end{subfigure}
~
    \begin{subfigure}[t]{0.32\textwidth}
        \centering
        \includegraphics[width=0.9\textwidth, trim={0 0 0 1.1cm}, clip]{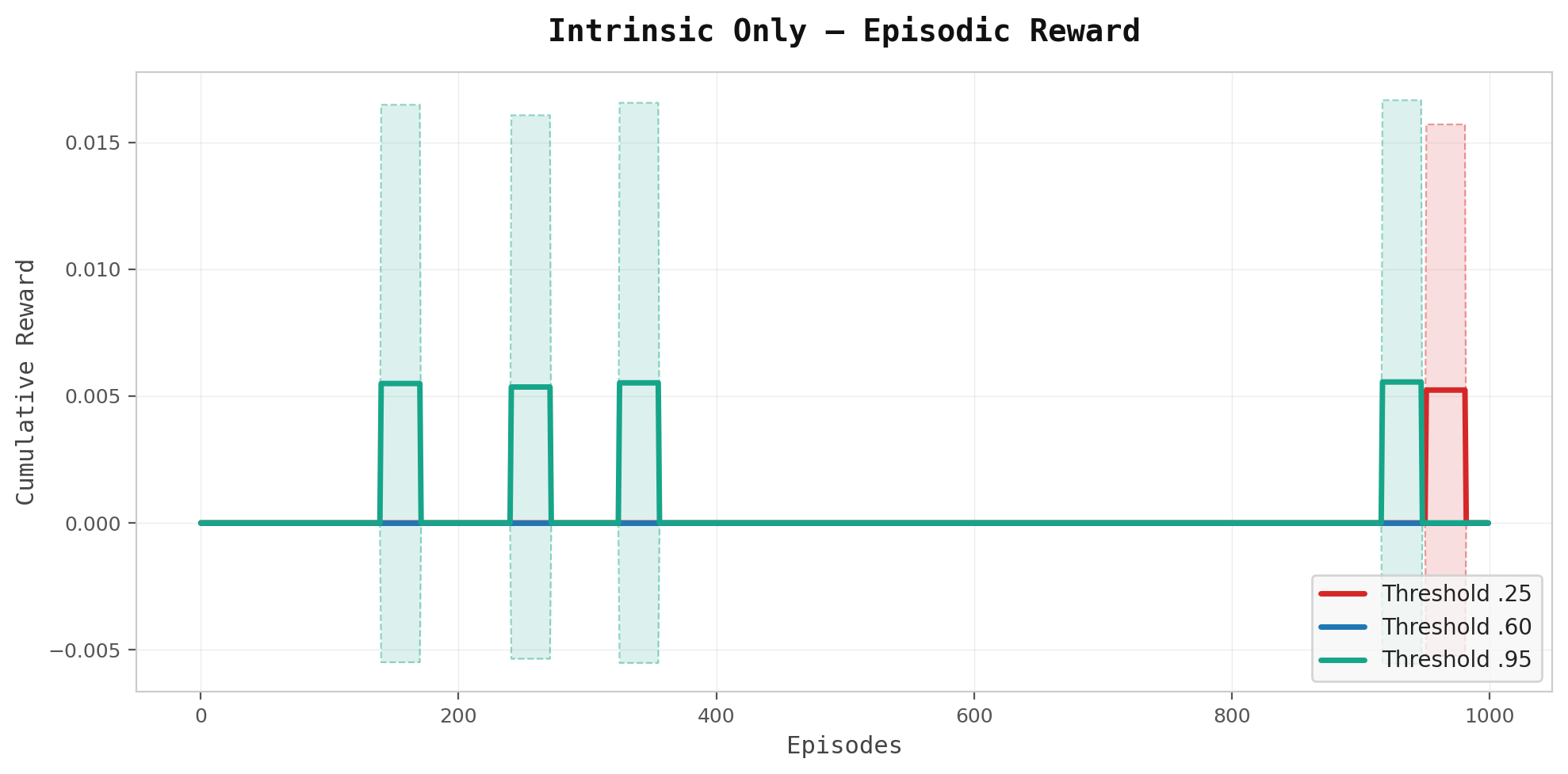}
        \caption{Episodic Reward}
        \label{fig:INT_EXTRINSIC}
    \end{subfigure}
    ~
    \begin{subfigure}[t]{0.32\textwidth}
        \centering
        \includegraphics[width=0.9\textwidth, trim={0 0 0 1.2cm}, clip]{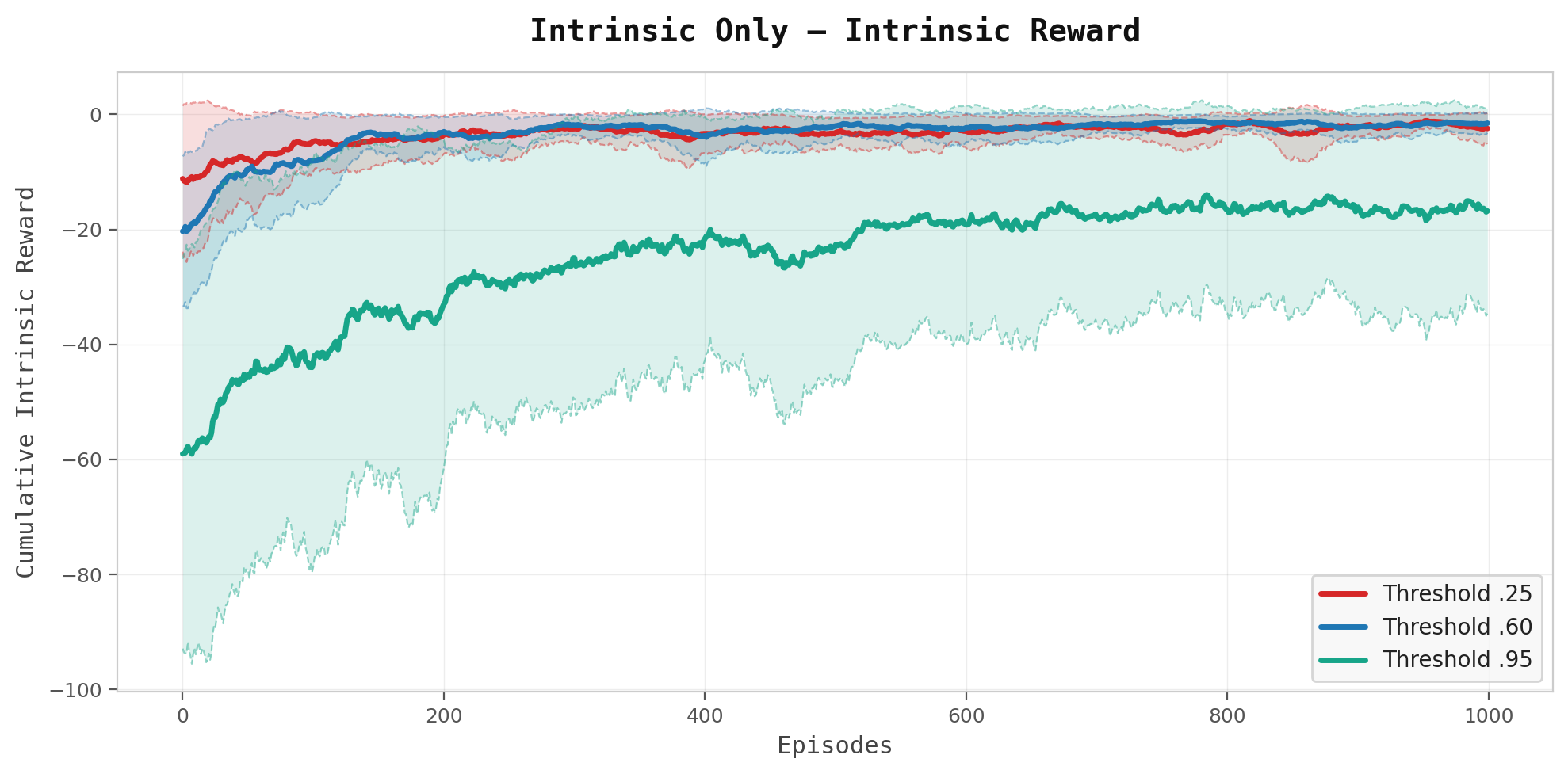}
        \caption{Episodic Intrinsic}
    \end{subfigure}

    \caption{Sidewalk intrinsic-only training curves across thresholds ($5$ runs, shaded variance). $\theta_{\text{thr}} = 0.60$ maintains survival above base PPO whereas $\theta_{\text{thr}} = 0.95$ declines below it.}
    \label{fig:Intrinsic_Only}
\vspace{-0.75cm}
\end{figure}

The low threshold ($\theta_{\text{thr}} = 0.25$) significantly outperforms the stimuli baseline (composite: $t(6.0) = 5.68$, $p = .001$, $d = 3.59$) but does not significantly outperform base PPO ($t(4.6) = 1.38$, $p = .232$), indicating that the classifier activates too infrequently to provide behavioral shaping beyond random exploration. High thresholds ($\theta_{\text{thr}} = 0.95$) produce bimodal outcomes: 
inspection of per-run data reveals that one or two of the five runs catastrophically fail while the others succeed, driving large standard deviations ($\pm 50.26$) and non-significant comparisons against base PPO. 
The large cumulative intrinsic penalties at high thresholds reflect a greater number of close approximations to the demonstrator's behavior, where closer approximations produce larger negative values. The intrinsic reward curve in Figure~\ref{fig:Intrinsic_Only}c illustrates this: 
the $\theta_{\text{thr}} = 0.95$ condition begins with cumulative penalties near $-60$ to $-95$ per episode, indicating that the strict threshold fires on nearly every observation early in training before the agent learns to avoid triggering states. These threshold sensitivity results connect to the inhibition paradigm of Section~\ref{subsec:reproduction}: too little gating produces a weak signal, while too much gating either fails to activate or overwhelms the gradient.

No significant differences were observed between composite and intrinsic-only conditions at any threshold (all $|t| < 1.45$, all $p > 0.21$), and extrinsic reward remained near zero across all conditions (Figure~\ref{fig:Intrinsic_Only}b), confirming that observed survival differences are driven entirely by the intrinsic vicarious signal. 
At $\theta_\text{thr} = 0.25$, the composite condition shows a marginal directional decrease relative to the intrinsic-only condition ($126.38$ vs.\ $130.59$), likely reflecting noise introduced into the PPO advantage estimate by the near-zero extrinsic reward. The agent does not solve the environment; however, it uncovers paths that prolong survival and occasionally stumbles into the goal.

\subsection{CarRacing Environment}
\label{subsec:car_racing_env}

\begin{table}[t]
\centering
\caption{CarRacing environment results over the final $200$ training episodes across $5$ runs (mean $\pm$ SD). VC: vicarious conditioning.}
\label{tab:carracing_result_table}
\resizebox{0.90\columnwidth}{!}{%
\begin{tabular}{lcccc}
\hline
Method & Episode Length & Ext Reward & Intr Reward & Composite Reward \\ \hline
\multicolumn{5}{c}{\textbf{Baselines}} \\ \hline
Base PPO & $33.27 \pm 2.37$ & $4.66 \pm 0.37$ & $0$ & $4.66 \pm 0.37$ \\
PPO + Stimuli~\citep{sanchez2024fear} & $24.38 \pm 6.79$ & $2.93 \pm 1.21$ & $-6.48 \pm 6.57$ & $-3.55 \pm 7.58$ \\ \hline
\multicolumn{5}{c}{\textbf{Positive VC (Approach)}} \\ \hline
PPO + VC ($\theta_{\text{thr}} = 0.20$) & $81.33 \pm 23.82$ & $1.65 \pm 1.27$ & $46.52 \pm 26.58$ & $48.17 \pm 25.59$ \\
PPO + VC ($\theta_{\text{thr}} = 0.60$) & $83.20 \pm 14.19$ & $2.05 \pm 1.03$ & $38.89 \pm 29.38$ & $40.94 \pm 28.48$ \\
PPO + VC ($\theta_{\text{thr}} = 0.90$) & $\mathbf{88.45 \pm 5.26}$ & $1.36 \pm 0.18$ & $46.52 \pm 11.91$ & $\mathbf{47.88 \pm 11.76}$ \\ \hline
\multicolumn{5}{c}{\textbf{Negative VC (Avoidance)}} \\ \hline
PPO + VC ($\theta_{\text{thr}} = 0.20$) & $24.28 \pm 5.74$ & $2.53 \pm 0.72$ & $-3.23 \pm 2.88$ & $-0.70 \pm 3.36$ \\
PPO + VC ($\theta_{\text{thr}} = 0.60$) & $18.43 \pm 1.78$ & $1.92 \pm 0.15$ & $-5.58 \pm 3.42$ & $-3.66 \pm 3.50$ \\
PPO + VC ($\theta_{\text{thr}} = 0.90$) & $24.79 \pm 4.71$ & $2.83 \pm 0.92$ & $-4.10 \pm 3.38$ & $-1.27 \pm 3.99$ \\ \hline
\multicolumn{5}{c}{\textbf{Composite VC (Positive + Negative)}} \\ \hline
PPO + VC ($\theta_{\text{thr}} = 0.20$) & $78.67 \pm 29.57$ & $1.09 \pm 0.39$ & $46.39 \pm 30.00$ & $47.48 \pm 29.63$ \\
PPO + VC ($\theta_{\text{thr}} = 0.60$) & $57.46 \pm 32.89$ & $1.82 \pm 0.73$ & $22.28 \pm 35.19$ & $24.10 \pm 34.60$ \\
PPO + VC ($\theta_{\text{thr}} = 0.90$) & $61.82 \pm 36.26$ & $1.51 \pm 0.32$ & $21.21 \pm 28.03$ & $22.72 \pm 27.81$ \\ \hline
\end{tabular}%
}
\end{table}

\begin{figure}[!h] 

   \centering
    \begin{subfigure}[t]{0.32\textwidth}
        \centering
        \includegraphics[width=0.9\textwidth, trim={0 0 0 1.2cm}, clip]{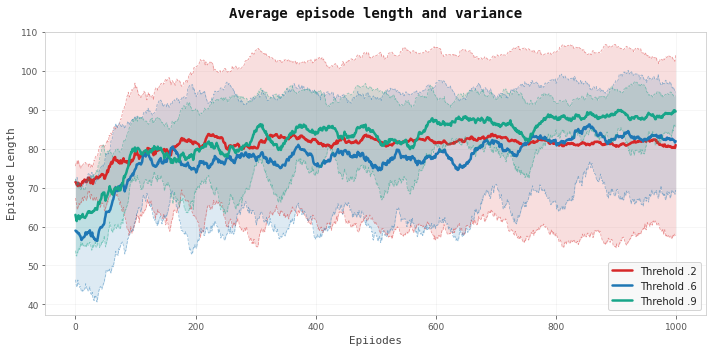}
        \caption{Episode Length}
        \label{fig:INT_LENGHT}
    \end{subfigure}
~
    \begin{subfigure}[t]{0.32\textwidth}
        \centering
       \includegraphics[width=0.9\textwidth, trim={0 0 0 1.1cm}, clip]{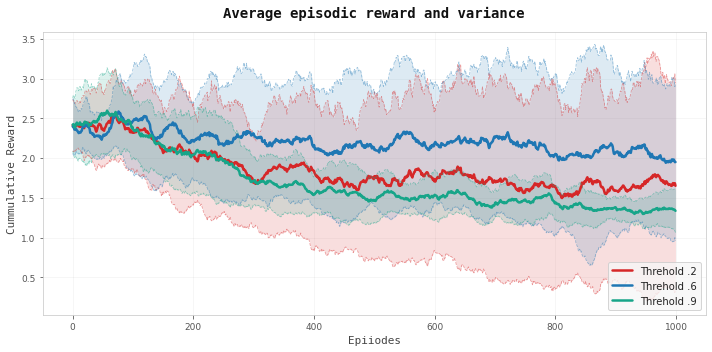}
        \caption{Episodic Reward}
        \label{fig:INT_EXTRINSIC}
    \end{subfigure}
    ~
    \begin{subfigure}[t]{0.32\textwidth}
        \centering
        \includegraphics[width=0.9\textwidth, trim={0 0 0 1.2cm}, clip]{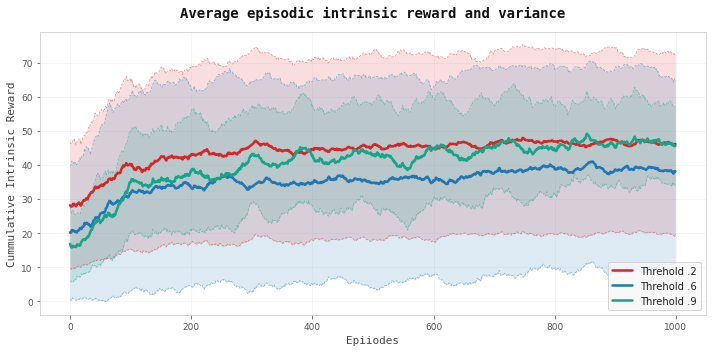}
        \caption{Episodic Intrinsic}
    \end{subfigure}

    \caption{CarRacing positive VC training curves across $\theta_{\text{thr}} \in \{0.2, 0.6, 0.9\}$ (5 runs, shaded variance). Positive VC increases episode length while reducing extrinsic reward, reflecting conservative center-lane driving over tile-visiting efficiency.}
    \label{fig:carracing_curves_main}

\end{figure}

The CarRacing environment tests positive VC (center-lane driving), negative VC (off-road driving), and composite VC (both VCs). All conditions use $26$ demonstrations per class / valence and $3$ for the trajectory length. A custom POMDP wrapper introduces a non-descriptive terminal condition when the agent contacts the grass boundary\footnote{See supplementary for a figure} and creates a field of view. The extrinsic reward is tile-based: 
the agent earns reward proportional to the number of track tiles visited per step. We test $\theta_{\text{thr}} \in \{0.20, 0.60, 0.90\}$ for each condition; training curves for positive VC appear in Figure~\ref{fig:carracing_curves_main}. 

Positive VC produces substantial improvements in episode length across all thresholds. The best-performing condition in the table ($\theta_{\text{thr}} = 0.9$) achieves $88.45$ ($95$\% CI $[81.92, 94.98]$) steps with low variance ($\pm 5.26$), nearly three times the episode length of base PPO at $33.27$ ($95$\% CI $[30.33, 36.21]$) ($t(5.6) = 19.13$, $p < .001$, $d = 13.53$). All positive VC thresholds significantly outperform base PPO ($\theta_{\text{thr}} = 0.6$: $83.20$ $[65.58, 100.82]$, $t(4.2) = 6.94$, $p = .002$, $d = 4.91$; $\theta_{\text{thr}} = 0.2$: $81.33$ $[51.75, 110.91]$, $t(4.1) = 4.02$, $p = .015$, $d = 2.85$). 
The method encourages the agent to replicate the demonstrated center-lane trajectory, resulting in conservative driving behavior. 

The positive VC agent earns significantly lower extrinsic reward than base PPO ($1.36 \pm 0.18$ vs.\ $4.66 \pm 0.37$ at $\theta_{\text{thr}} = 0.9$; $t(5.8) = -17.93$, $p < .001$). This tradeoff reflects the reward structure rather than a failure of the method. The tile-based reward favors aggressive driving that visits more tiles per step; base PPO optimizes this reward rate ($0.14$ reward/step) but terminates quickly upon contacting the grass boundary. The VC agent, guided by the center-lane demonstration, takes a geometrically longer path per tile ($0.02$ reward / step) but survives nearly three times longer. The agent is doing exactly what the demonstration taught: it prioritizes the center of the lane over tile-visiting efficiency. The composite reward (extrinsic + intrinsic) captures the full optimization objective, where positive VC at $\theta_{\text{thr}} = 0.9$ achieves $47.88 \pm 11.76$ compared to base PPO's $4.66 \pm 0.37$.

Negative VC performs at or below base PPO across all thresholds ($\theta_{\text{thr}} = 0.6$: $18.43$ $[16.22, 20.64]$; $t(7.4) = -11.20$, $p < .001$, $d = -7.08$), confirming that avoidance conditioning is inappropriate for an approach task. The negative demonstration penalizes the agent for proximity to the track boundary, but in the CarRacing environment, the agent must stay \textit{on} the track to survive; the avoidance signal conflicts with the task structure and produces performance indistinguishable from the stimuli baseline ($\theta_{\text{thr}} = 0.2$ vs.\ Stimuli: $t(7.8) = -0.02$, $p = .988$). This result validates that the method is valence-sensitive: the sign of $v_\tau$ determines whether VC helps or hinders, consistent with the biological distinction between fear conditioning and appetitive conditioning.

Composite VC shows mixed results. At $\theta_{\text{thr}} = 0.2$, composite VC achieves $78.67$ $[41.96, 115.38]$ steps, still significantly above the baseline PPO. 
At higher thresholds ($\theta_{\text{thr}} \geq 0.6$), composite VC becomes unstable. The conflicting valence signals at strict thresholds appear to destabilize the policy gradient, suggesting that multi-valence conditioning requires careful threshold calibration.

\section{Discussion and Conclusion}
\label{sec:discussion}

Throughout our experiments, we test vicarious conditioning's ability to alter the agent's behavior to match that of a demonstrator. Crucially, our inclusion of non-descriptive terminal conditions in both environments alter the reward landscape. CarRacing's native step-cost incentives aggressive driving that visits tiles quickly, which conflicts with the safe center-lane behavior demonstrated. This results in a lower extrinsic reward, but nearly threefold longer survival. Conversely, Sidewalk provides no gradient for the non-descriptive terminal condition, but VC supplies the missing avoidance signal permitting the agent to explore long enough to occasionally reach the goal.

Vicarious conditioning (VC) demonstrated behavior that dominates the visual field -- the sidewalk boundary in Sidewalk, the track center in CarRacing -- producing the strongest intrinsic signal. CarRacing further demonstrates that VC can learn from demonstrations with variable, non-equivalent actions; confirming the method's action-invariance property. The variance in episode length across vicarious conditions and thresholds reveals a symmetry: a low threshold for negative VC produces the same pattern as a high threshold for positive VC. Both VCs enforce equivalent strictness towards the demonstrated ``safe'' behavior. The composite reward exhibited high variance across both environments, since the agent must balance the extrinsic and intrinsic rewards concurrently. Two reward functions are conditionally present; thus, the agent's explorations leads to performance variation.

Our work produces a vicarious conditioning method that uses memory for low-shot reproduction of demonstrator behavior. This algorithm opens the door to deploying robotic systems with minimal behavioral representations, even across heterogeneous action spaces. Furthermore, the inhibition threshold provides a tunable risk parameter: 
high-risk behaviors require higher classifier confidence before the intrinsic reward activates. Two limitations exist due to the incorporation of the MANN. First, the MANN architecture is a fundamental constraint: 
behaviors that are indistinguishable to the feature extractor cannot be separated in memory. Second, our method does not permit memory extinction; once behavior is stored, its vicarious value persists even if the environment contradicts it. The reinforcement step 
will eventually ensure the extrinsic reward dominates, but the intrinsic conditioned fear persists, leading to a `done' condition, which impacts the PPO advantage calculation.

This work addressed the gap between biological agents' ability to learn from demonstrations and RL's reliance on direct experience by introducing intrinsic VC. We provided a formal framework grounded in the four biological vicarious conditioning steps: 
attention, retention, reproduction, and reinforcement; 
which was implemented using a MANN architecture. 
The sidewalk results demonstrate that negative VC enables avoidance of non-descriptive terminal conditions, yielding longer episodes driven by the intrinsic signal. The CarRacing results demonstrate that positive vicarious conditioning produces conservative driving with longer survival and composite reward  whereas negative conditioning correctly fails in the approach task. Overall, vicarious conditioning enables RL agents to learn from few observed demonstrations without explicit policies, large datasets, or knowledge of the demonstrators reward function. We believe that this form of conditioning will play a critical role in developing capable, single-life RL agents.

\bibliography{main}
\bibliographystyle{rlj}

\beginSupplementaryMaterials

\setcounter{secnumdepth}{0}

\section{Vicarious Conditioning Outline}
\begin{figure}[!h]
     \centering
     \includegraphics[width=0.9\textwidth, trim={0 0 0 0.7cm}, clip]{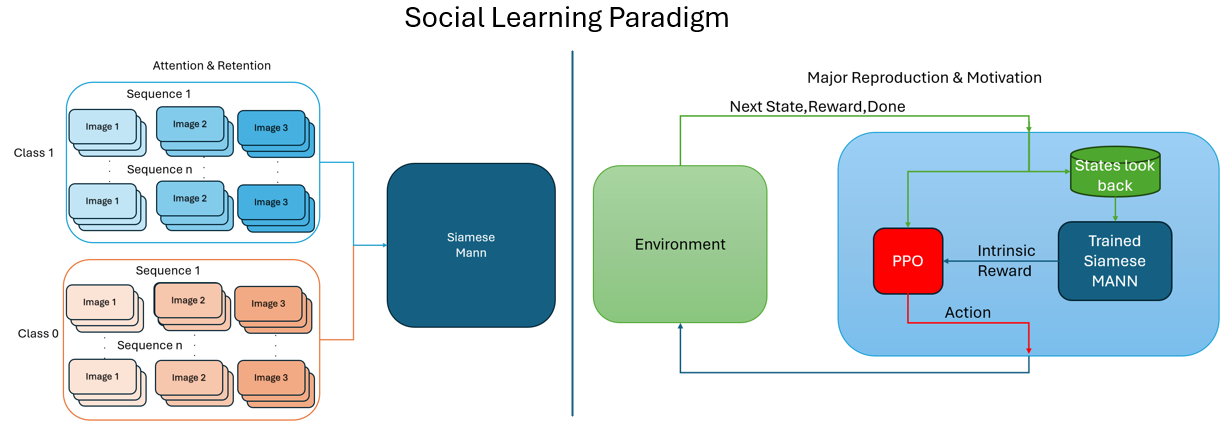} 
     \caption{Depicts the social learning frameworks where attention and retention use low-shot learning to train the Siamese MANN while reproduction and motivation occur through intrinsic rewards.}
     \label{fig:total_method}
\end{figure}

The biological vicarious conditioning process uses four distinct steps: attention, retention, reproduction, and reinforcement. Our work splits the method into two processes to facilitate the use of neural architectures. We depict the delineation into two distinct learning processes \ref{fig:total_method}.

\begin{figure}[h]
  \centering
  \includegraphics[width=\textwidth, trim={0 0 0 0.5cm}, clip]{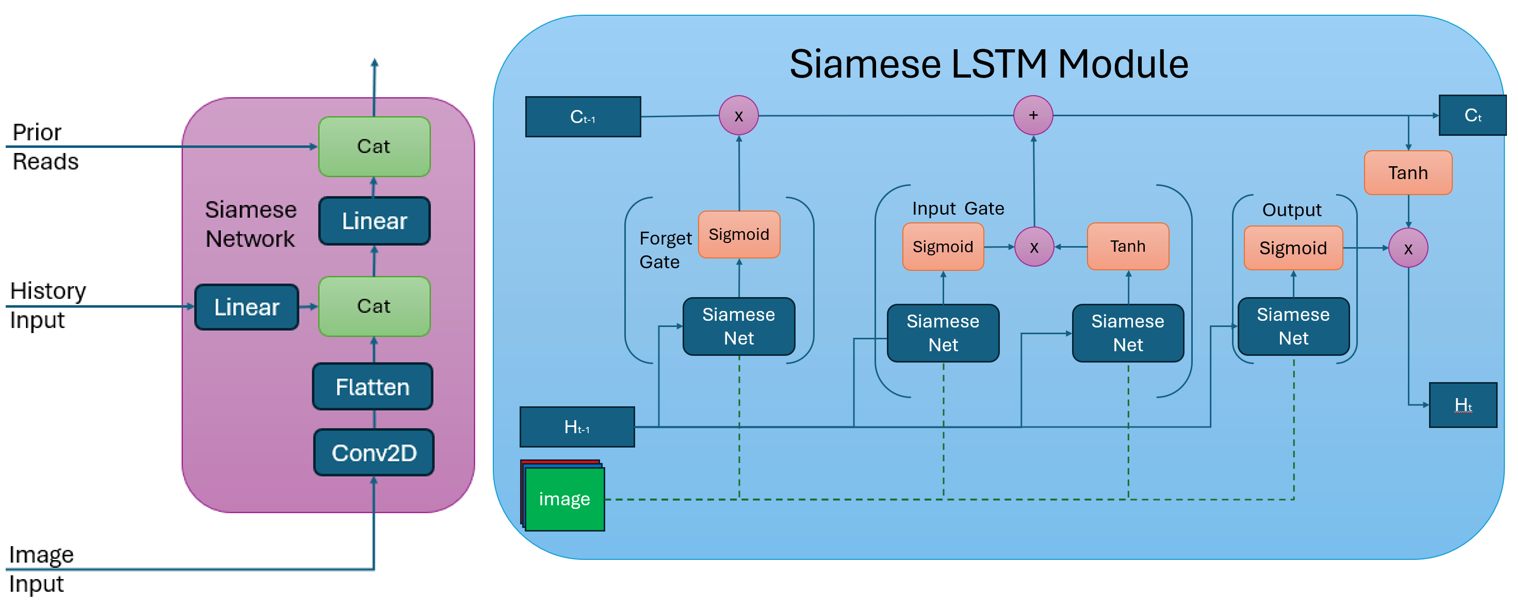}
  \caption{Demonstrates the Siamese Network that is then used as a gate for the SIAMESE LSTM Module, the inclusion of this network for the gates allows for the mixing of images and vectors}
  \label{fig:SLSTM_CELL_FULL}
\end{figure}

\section{Siamese LSTM}

Our siamese LSTM replaces each LSTM gate with a siamese network, allowing the network to process a sequence of images, the previous memory read values, and the cell state. Figure \ref{fig:SLSTM_CELL_FULL} provides a succinct overview of the major changes to the layer. We provide the mathematical equivalent of the encoding process in equation \ref{eq:controller_en}. Finally, we provide the complete pseudocode in algorithms \ref{Psuedo_Code_2} and \ref{Psuedo_Code_1}. Algorithm \ref{Psuedo_Code_2} depicts the Attention and Retention, which we implemented through our MANN architecture. Then reproduction and reinforcement are depicted in pseudo-code \ref{Psuedo_Code_1} which demonstrates the read function of the Mann architecture and its effect on PPO's training.

\begin{equation}
\begin{aligned}
\mathcal{G}&= \sigma(W_{l\cdots l-1} [ \mathbf{State}_{l\cdots l_{n}},h_{t-1},\mathbf{r}_t ] + b_{l\cdots l_{n}})\\
f_t &= \sigma\left(\mathcal{G}_f ([\mathbf{State}_{l\cdots l_{n}},h_{t-1},\mathbf{r}_t] )\right) \\
i_t &= \sigma\left(\mathcal{G}_i ([\mathbf{State}_{l\cdots l_{n}},h_{t-1},\mathbf{r}_t] )\right) \\
\tilde{C}_t &= \tanh\left(\mathcal{G}_C ([\mathbf{State}_{l\cdots l_{n}},h_{t-1},\mathbf{r}_t] )\right) \\
C_t &= f_t \odot \mathcal{C}_{t-1} + i_t \odot \tilde{C}_t \\
\mathbf{o}_t &= \sigma\left(\mathcal{G}_o([\mathbf{State}_{l\cdots l_{n}},h_{t-1},\mathbf{r}_t] )\right) \\
\mathbf{h}_t &= o_t \odot \tanh(C_t)
\end{aligned}
\label{eq:controller_en}
\end{equation}

\section{Experimental Setup}
\textbf{Wrappers and Dataset}

\begin{figure}[http]
    \medskip
    \centering
    \begin{subfigure}[b]{0.42\textwidth}
    \includegraphics[trim={0 0 0 0}, clip, width=\textwidth]{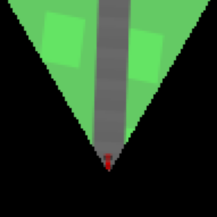}
    \caption{$\Delta-\forall$: State}
    \label{fig:POMDP}
    \end{subfigure}
    \hfill
    \begin{subfigure}[b]{0.43\textwidth}
    \includegraphics[trim={0 0 0 1 cm}, clip, width=\textwidth]{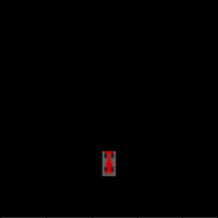}
    \caption{Collision Region}
    \label{fig:CollisionRegion}
    \end{subfigure}
    \caption{Collectively demonstrates the POMDP wrapper and region that considers collision. CarRacing maintains the car center to the environment; therefore, we checked the cropped region described in figure \ref{fig:CollisionRegion}. The POMDP cone provides a limited frame that rotates with the agent's viewpoint (the agent is stationary relative to the frame).}
    \label{fig:NDTCPOMDP_Wrapper}
\end{figure}
\begin{figure}[h!]
\centering
    \begin{subfigure}[t]{0.5\textwidth}
        \centering
        \includegraphics[width=0.9\textwidth, trim={0 0 0 1.2cm}, clip]{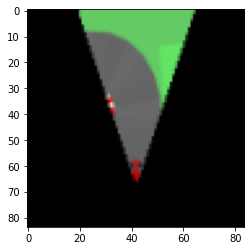}
        \caption{Behavior step 1}
        \label{fig:avoid_step1}
    \end{subfigure}
\hfill
    \begin{subfigure}[t]{0.5\textwidth}
        \centering
        \includegraphics[width=0.9\textwidth, trim={0 0 0 1.1cm}, clip]{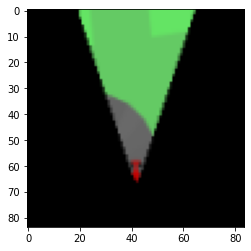}
        \caption{Behavior step 2}
        \label{fig:avoid_step2}
    \end{subfigure}

\vskip\baselineskip
    \begin{subfigure}[t]{0.5\textwidth}
        \centering
        \includegraphics[width=0.9\textwidth, trim={0 0 0 1.2cm}, clip]{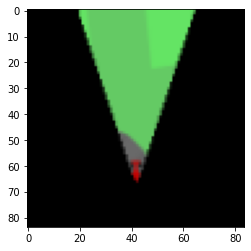}
        \caption{Behavior step 3}
        \label{fig:avoid_step3}
    \end{subfigure}
\caption{Describes an example of avoindance behavior for vicarious conditioning where the agent observes the demonstrator turn torwards the grass. Importantly, the steps in between represented behaviors have intermittent, non-represented steps (braking or accelerating) to reach the state; collectively, the agent is only provided the sequence with a known value.}
\label{fig:Avoid Class}
\end{figure}
\begin{figure}[h!]
\centering
    \begin{subfigure}[t]{0.5\textwidth}
        \centering
        \includegraphics[width=0.9\textwidth, trim={0 0 0 1.2cm}, clip]{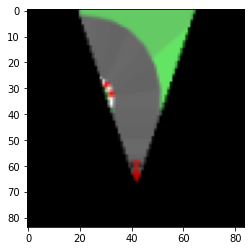}
        \caption{Behavior step 1}
        \label{fig:good_step1}
    \end{subfigure}
\hfill
    \begin{subfigure}[t]{0.5\textwidth}
        \centering
        \includegraphics[width=0.9\textwidth, trim={0 0 0 1.1cm}, clip]{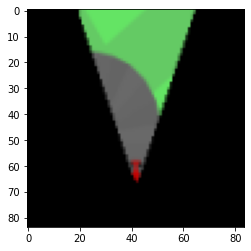}
        \caption{Behavior step 2}
        \label{fig:good_step2}
    \end{subfigure}

\vskip\baselineskip
    \begin{subfigure}[t]{0.5\textwidth}
        \centering
        \includegraphics[width=0.9\textwidth, trim={0 0 0 1.2cm}, clip]{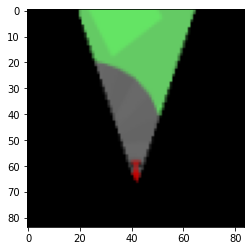}
        \caption{Behavior step 3}
        \label{fig:good_step3}
    \end{subfigure}
\caption{Describes an example of enticement behavior for vicarious conditioning where the agent learns to stay in the center of the lanes. Importantly, the steps in between represented behaviors have intermittent, non-represented steps (braking or accelerating) to reach the state; collectively, the agent is only provided the sequence with a known value.}
\label{fig:Good Class}
\end{figure}

To extend CarRacing to work under the same conditions as Sidewalk, we generated a wrapper that produces a POMDP and a non-descriptive terminal condition. Cohesively, we show the wrapper in Figure \ref{fig:NDTCPOMDP_Wrapper}, with Figure \ref{fig:POMDP} showing an example of the agents' state space. Figure \ref{fig:CollisionRegion} provides the boundary of the collision region for the agent. We applied the wrapper to both parts of the vicarious process, including dataset generation.  We provide an example of a positive behavior in Figure \ref{fig:Good Class} and a negative behavior in Figure \ref{fig:Avoid Class}.

\textbf{Run and Agent Parameters}

\begin{algorithm}[t]
\caption{Attention and Retention: SMANN Training}
\label{Psuedo_Code_2}
\begin{algorithmic}[1]
    \Procedure{TrainSMANN}{$\mathcal{D}$, $n$}
        \State \textbf{Input:} Demonstration dataset $\mathcal{D} = \{(\tau_i, v_{\tau_i})\}_{i=1}^{N}$, epochs $n$
        \State Initialize Siamese controller (S-LSTM), read/write heads, memory $\mathbf{M}$, output layer $(\mathbf{W}, \mathbf{b})$
        \State Zero-pad initial controller state, read vectors, and head state
        \For{epoch $= 1$ \textbf{to} $n$}
            \For{each batch $(\tau, v_\tau)$ in $\mathcal{D}$}
                \State $\mathbf{e}_\tau \leftarrow \text{S-LSTM}(\tau,\, h_{t-1},\, \mathbf{r}_t)$ \Comment{Controller encoding}
                \State $\mathbf{r}_t \leftarrow \text{ReadHeads}(\mathbf{e}_\tau, \mathbf{M})$ \Comment{Content-based read}
                \State $\mathbf{M} \leftarrow \text{WriteHeads}(\mathbf{e}_\tau, \mathbf{M})$ \Comment{LRU write}
                \State $\mathbf{p} \leftarrow \text{softmax}(\mathbf{W}[\mathbf{e}_\tau;\, \mathbf{r}_t] + \mathbf{b})$
                \State $\omega \leftarrow \text{sign}(v_\tau)$ \Comment{Discretize to $\omega \in \Omega$}
                \State $\mathcal{L} \leftarrow \text{CrossEntropy}(\mathbf{p},\, \omega)$
                \State Update $(\text{S-LSTM},\, \mathbf{W},\, \mathbf{b})$ via gradient descent on $\mathcal{L}$
            \EndFor
        \EndFor
        \State \textbf{Freeze} all SMANN parameters
    \EndProcedure
\end{algorithmic}
\end{algorithm}

\begin{algorithm}[t]
\caption{Reproduction and Reinforcement: PPO Training with Frozen SMANN}
\label{Psuedo_Code_1}
\begin{algorithmic}[1]
    \Procedure{TrainPPO}{frozen SMANN, $\alpha$, $\theta_{\text{thr}}$}
        \State Initialize environment, PPO policy $\pi$, replay buffer $\mathcal{B}$, observation buffer $\mathbf{B}_\tau$ of length $l$
        \For{episode $= 1$ \textbf{to} $1000$}
            \State Reset environment; $o_0 \leftarrow$ initial observation; done $\leftarrow$ False
            \While{\textbf{not} done}
                \State $a_t \sim \pi(o_t)$
                \State $o_{t+1},\, r^{\text{ext}}_t,\, \text{done} \leftarrow \text{env.step}(a_t)$
                \State Append $o_{t+1}$ to $\mathbf{B}_\tau$
                \State $\mathbf{p}_t \leftarrow \text{SMANN}(\mathbf{B}_\tau)$ \Comment{Frozen forward pass}
                \State $r^{\text{vic}}_t \leftarrow \sum_{\omega \in \Omega} \mathbf{1}[p_{t,\omega} > \theta_{\text{thr}}] \cdot v_{\tau,\omega} \cdot p_{t,\omega}$
                \State $\bar{r}_t \leftarrow r^{\text{ext}}_t + \alpha \cdot r^{\text{vic}}_t$
                \State $\mathcal{B}.\text{append}(o_t,\, o_{t+1},\, \bar{r}_t,\, \text{done})$
                \If{$|\mathcal{B}| \geq$ \textit{update\_steps}}
                    \State Update $\pi$ via PPO on $\mathcal{B}$; clear $\mathcal{B}$
                \EndIf
            \EndWhile
        \EndFor
    \EndProcedure
\end{algorithmic}
\end{algorithm}

We provide the hyperparameters used across both MANN and reinforcement learning below in Tables \ref{tab:ENV_Hyper} and \ref{tab:Intrinsic_Hyper} to promote reproducibility. Both tables provide the hyperparameters used per run and per environment. 

\begin{table}[]
\begin{center}
\caption{This table provides the hyperparameter for all runs performed across all tested environments. A crucial difference is the max step limit in each environment, which limits the maximum achievable number of steps: 150 for sidewalk and 350 for CarRacing. Furthermore, we include the agent training hyperparameters, such as the update rate and the number of epochs per episode update, for PPO.}

\label{tab:ENV_Hyper}
\begin{tabular}{lcllll}
\hline
\multicolumn{6}{c}{\textbf{Environment and Agent Run Hyper Parameters}}                                                                                        \\ \hline \\ 
\multicolumn{1}{c}{\textbf{Hyperparameters}}        & \multicolumn{5}{c}{\textbf{CarRacing with Wrapper}}                                                       \\  \\ \hline \\
\multicolumn{1}{c}{Clip Coefficient}       & \multicolumn{5}{c}{0.2}                                                                       \\ 
\multicolumn{1}{c}{Total Episodes}           & \multicolumn{5}{c}{1000}                                                                   \\ 
\multicolumn{1}{c}{Max Steps}           & \multicolumn{5}{c}{350}                                                                   \\ 
\multicolumn{1}{c}{Gamma}                 & \multicolumn{5}{c}{.99}                                                                       \\ 
\multicolumn{1}{c}{Learning Rate}         & \multicolumn{5}{c}{1e-5}                                                                      \\ 
\multicolumn{1}{l}{Policy Update Steps}   & \multicolumn{5}{c}{400}                                                                       \\ 
\multicolumn{1}{l}{Update Epoch}          & \multicolumn{5}{c}{60}    
            \\ 
\multicolumn{1}{c}{parallel environments} & \multicolumn{5}{c}{1}        
            
            \\ \hline\\
            
\multicolumn{1}{c}{\textbf{Hyperparameters}}        & \multicolumn{5}{c}{\textbf{
Sidewalk}}                                                      \\ \\ \hline\\ 
\multicolumn{1}{c}{Clip Coefficient}       & \multicolumn{5}{c}{0.2}                                                                       \\
\multicolumn{1}{c}{Total Episodes}           & \multicolumn{5}{c}{1000}                                                                   \\ 
\multicolumn{1}{c}{Max Steps}           & \multicolumn{5}{c}{150}   \\
\multicolumn{1}{c}{Gamma}                 & \multicolumn{5}{c}{.99}                                                                       \\ 
\multicolumn{1}{c}{Learning Rate}         & \multicolumn{5}{c}{1e-5}                                                                      \\ 
\multicolumn{1}{l}{Policy Update Steps}   & \multicolumn{5}{c}{400}                                                                       \\ 
\multicolumn{1}{l}{Update Epoch}          & \multicolumn{5}{c}{60}    
            \\ 
\multicolumn{1}{c}{parallel environments} & \multicolumn{5}{c}{1}        
            
            \\ 
\end{tabular}
\end{center}
\end{table}

\begin{table}[]
\begin{center}
\caption{This table describes the chosen hyperparameters for the intrinsic rewards we tested. Crucially, for stimulus fear, we resize the input images to a black-and-white 40x40 image. We reduced the image size to allow a feasible number of neurons on the input layer. Finally, all other hyperparameters are fixed across each method (besides controller architecture), and the CarRacing environment posed a larger amount of representations to learn, requiring a larger amount of memory and read and write heads for both methods.}

\label{tab:Intrinsic_Hyper}

\begin{tabular}{lcllll}
\hline
\multicolumn{6}{c}{\textbf{Per Environment Intrinsic Reward Hyper Parameters}}                                                                                        \\ \hline \\
\multicolumn{1}{c}{\textbf{Hyperparameters}}        & \multicolumn{5}{c}{\textbf{CarRacing with Wrapper}}                                                       \\ \\ \hline
\multicolumn{1}{c}{Controller Depth}       & \multicolumn{5}{c}{7}                                                                       \\ 
\multicolumn{1}{c}{Num Memory Size}           & \multicolumn{5}{c}{N= 128, M=60}                                                                   \\ 
\multicolumn{1}{c}{Num Read Heads}           & \multicolumn{5}{c}{30}                                                                   \\ 
\multicolumn{1}{c}{Num Write Heads}          & \multicolumn{5}{c}{30}                                                                       \\ 
\multicolumn{1}{c}{Learning Rate}         & \multicolumn{5}{c}{7e-5}                                                                      \\ 
\multicolumn{1}{c}{Image Dimensions}         & \multicolumn{5}{c}{(3,84,84)}                                                                      \\ 
\multicolumn{1}{c}{Num Examples}         & \multicolumn{5}{c}{26 Per Class}                                                                      \\ 
\multicolumn{1}{l}{Training Epochs}   & \multicolumn{5}{c}{300}                                                                      
            \\ \hline\\
            
\multicolumn{1}{c}{\textbf{Hyperparameters}}        & \multicolumn{5}{c}{\textbf{
Sidewalk}}                                                      \\ \\ \hline\\ 
\multicolumn{1}{c}{Controller Depth}       & \multicolumn{5}{c}{7}                                                                       \\ 
\multicolumn{1}{c}{Num Memory Size}           & \multicolumn{5}{c}{N= 128, M=40}                                                                   \\ 
\multicolumn{1}{c}{Num Read Heads}           & \multicolumn{5}{c}{10}                                                                   \\ 
\multicolumn{1}{c}{Num Write Heads}          & \multicolumn{5}{c}{10}                                                                       \\ 
\multicolumn{1}{c}{Learning Rate}         & \multicolumn{5}{c}{1e-5}                                                                      \\ 
\multicolumn{1}{c}{Image Dimensions}         & \multicolumn{5}{c}{(3,40,40)}                                                                      \\ 
\multicolumn{1}{c}{Num Examples}         & \multicolumn{5}{c}{26 Per Class}                                                                      \\ 
\multicolumn{1}{l}{Training Epochs}   & \multicolumn{5}{c}{150}                                                                      
            \\ \hline\\ \\

\end{tabular}
\end{center}
\end{table}

\section{Non-Central Supplemental Results}
\begin{figure}[h!]
\centering
    \begin{subfigure}[t]{0.5\textwidth}
        \centering
        \includegraphics[width=0.9\textwidth, trim={0 0 0 1.2cm}, clip]{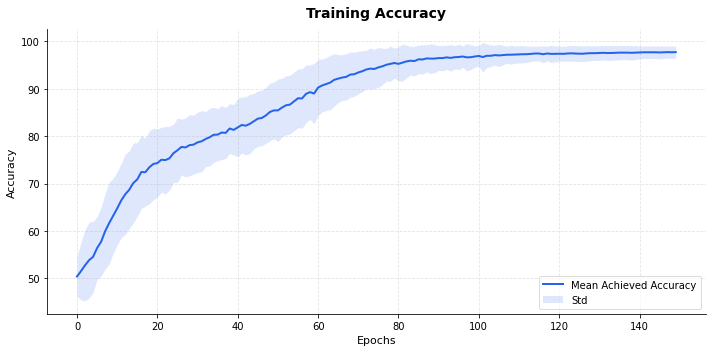}
        \caption{SideWalk}
        \label{fig:SideWalk_acc}
    \end{subfigure}
\hfill
    \begin{subfigure}[t]{0.5\textwidth}
        \centering
        \includegraphics[width=0.9\textwidth, trim={0 0 0 1.1cm}, clip]{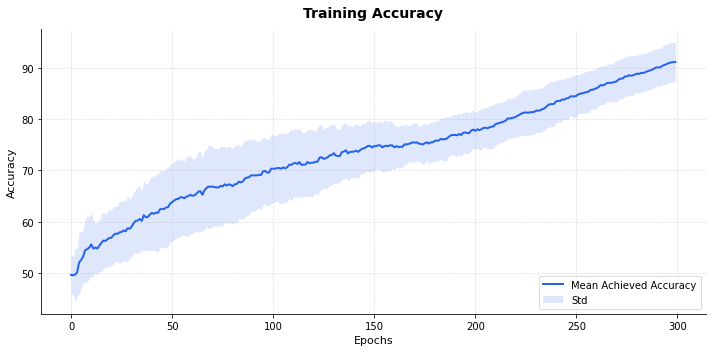}
        \caption{CarRacing}
        \label{fig:CarRacing_acc}
    \end{subfigure}

\caption{Demonstrates the achieved average and standard deviation of the accuracies for SMANN training. Both methods achieved above 90 percent accuracy with CarRacing showing lower accuracies due to greater complexity in representation).}
\label{fig:Mann_accuracies}
\end{figure}

\begin{figure}[!h] 

   \centering
    \begin{subfigure}[t]{0.8\textwidth}
        \centering
        \includegraphics[width=0.9\textwidth, trim={0 0 0 1.2cm}, clip]{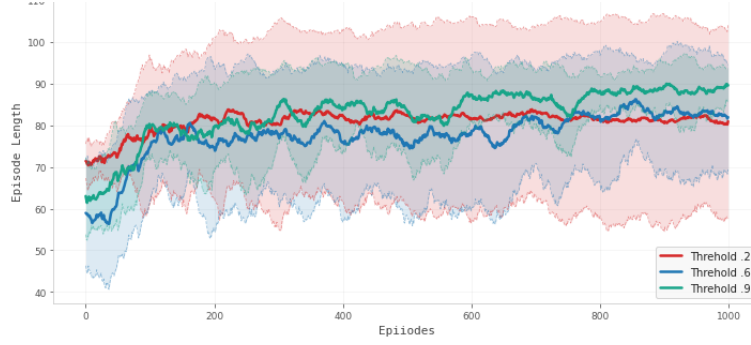}
        \caption{Episode Length}
        \label{fig:INT_LENGHT}
    \end{subfigure}
~
    \begin{subfigure}[t]{0.8\textwidth}
        \centering
       \includegraphics[width=0.9\textwidth, trim={0 0 0 1.1cm}, clip]{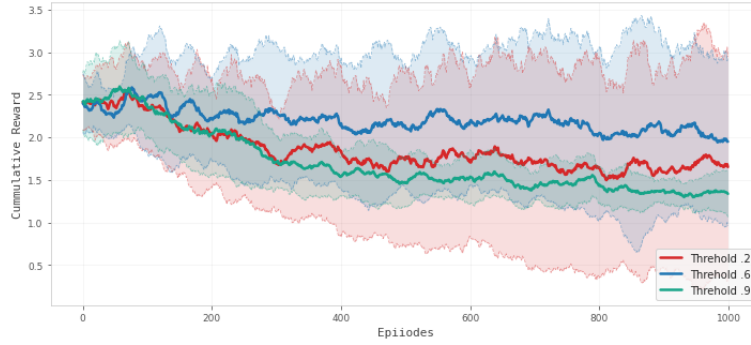}
        \caption{Episodic Reward}
        \label{fig:INT_EXTRINSIC}
    \end{subfigure}
    ~
    \begin{subfigure}[t]{0.8\textwidth}
        \centering
        \includegraphics[width=0.9\textwidth, trim={0 0 0 1.2cm}, clip]{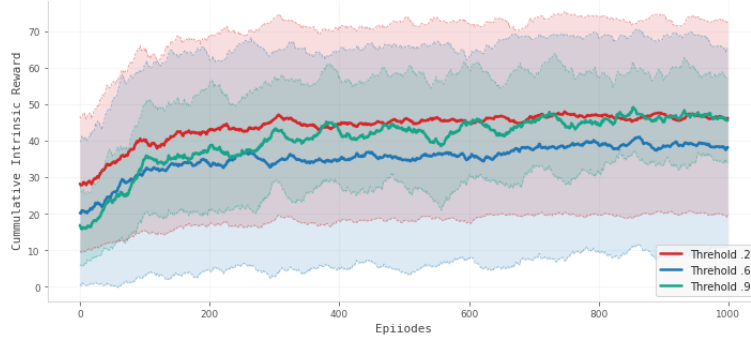}
        \caption{Episodic Intrinsic}
    \end{subfigure}

    \caption{Training curves for the positive Vicarious conditioning with extrinsic rewards in the CarRacing environment across three representative thresholds ($\theta_{\text{thr}} \in \{0.2, 0.6, 0.9\}$), averaged over 5 runs with variance shown as shaded regions. The results show that positive intrinsic reward promotes an increase in episode length. Importantly the extrinsic is lower since the agent is not just optimizing the tile visitation per action (extrinsic reward).}
    \label{fig:carracing_curves_positve}

\end{figure}

\begin{figure}[!h] 

   \centering
    \begin{subfigure}[t]{0.8\textwidth}
        \centering
        \includegraphics[width=0.9\textwidth, trim={0 0 0 1.2cm}, clip]{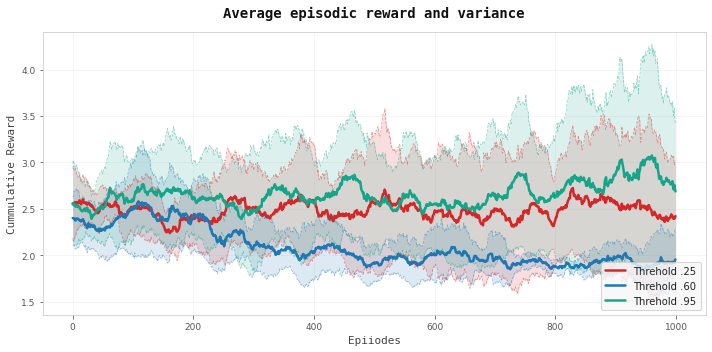}
        \caption{Episode Length}
        \label{fig:INT_LENGHT}
    \end{subfigure}
~
    \begin{subfigure}[t]{0.8\textwidth}
        \centering
       \includegraphics[width=0.9\textwidth, trim={0 0 0 1.1cm}, clip]{Media/Results/CarRacing/VC_Negative_n_Extrinsic/episodic_extrinsic_reward.png}
        \caption{Episodic Reward}
        \label{fig:INT_EXTRINSIC}
    \end{subfigure}
    ~
    \begin{subfigure}[t]{0.8\textwidth}
        \centering
        \includegraphics[width=0.9\textwidth, trim={0 0 0 1.2cm}, clip]{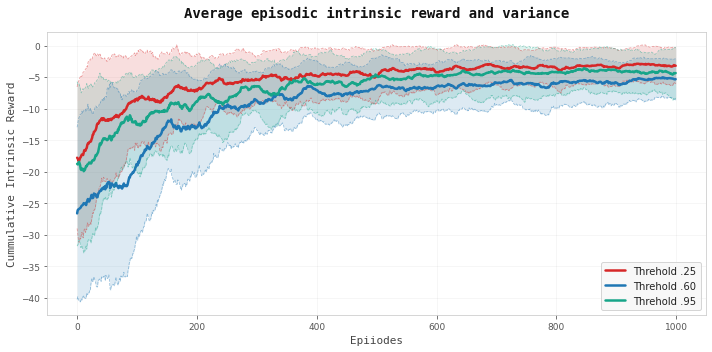}
        \caption{Episodic Intrinsic}
    \end{subfigure}

    \caption{Training curves for the negative Vicarious conditioning with extrinsic rewards in the CarRacing environment across three representative thresholds ($\theta_{\text{thr}} \in \{0.2, 0.6, 0.9\}$), averaged over 5 runs with variance shown as shaded regions. The graphs demonstrate the intrinsic reward was optimized but that only the high threshold pushed for substantial increase in the cumulative extrinsic reward.}
    \label{fig:carracing_curves_negative}

\end{figure}

\begin{figure}[!h] 

   \centering
    \begin{subfigure}[t]{0.8\textwidth}
        \centering
        \includegraphics[width=0.9\textwidth, trim={0 0 0 1.2cm}, clip]{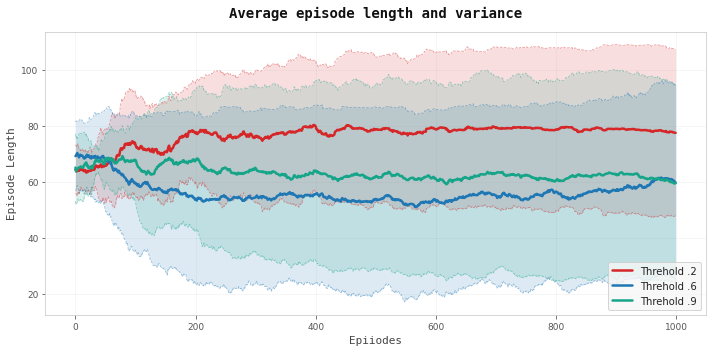}
        \caption{Episode Length}
        \label{fig:INT_LENGHT}
    \end{subfigure}
~
    \begin{subfigure}[t]{0.8\textwidth}
        \centering
       \includegraphics[width=0.9\textwidth, trim={0 0 0 1.1cm}, clip]{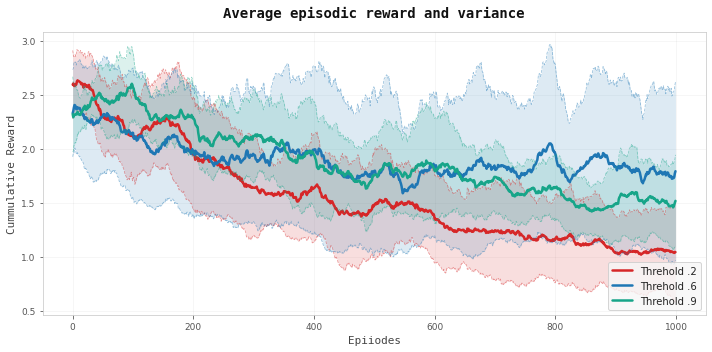}
        \caption{Episodic Reward}
        \label{fig:INT_EXTRINSIC}
    \end{subfigure}
    ~
    \begin{subfigure}[t]{0.8\textwidth}
        \centering
        \includegraphics[width=0.9\textwidth, trim={0 0 0 1.2cm}, clip]{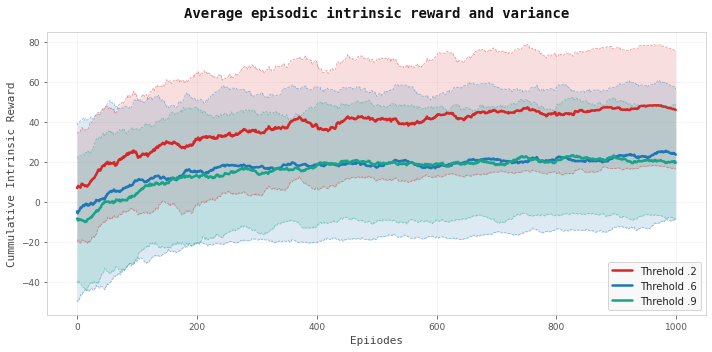}
        \caption{Episodic Intrinsic}
    \end{subfigure}

    \caption{Training curves for the Composite Vicarious conditioning with extrinsic rewards in the CarRacing environment across three representative thresholds ($\theta_{\text{thr}} \in \{0.2, 0.6, 0.9\}$), averaged over 5 runs with variance shown as shaded regions. The results show greater variance in the achieved episode length which is byproduct of optimizing two conditional reward functions and the extrinsic reward.}
    \label{fig:carracing_curves_composite}

\end{figure}

Finally, in this section, we provided a more extensive version of the prior results. First, we provided the average achieved accuracy across all runs per environment for our SMANN architecture in Figure \ref{fig:Mann_accuracies}. Where Figure \ref{fig:SideWalk_acc} demonstrates the achieved accuracy in the sidewalk environment and \ref{fig:CarRacing_acc} the achieved accuracy in the CarRacing environment. Caracing shows lower accuracy in environments with greater complexity. Continuing, we demonstrate the previous intrinsic-only negative reward on the sidewalk in figure \ref{fig:sidewalk_Intrinsic_Only} and the composite vicarious conditioning values in figure \ref{fig:carracing_curves_negative}. Finally, we demonstrate all cumulative rewards, episode length and intrinsic rewards for all variations in Vicarious conditioning in CarRacing in figures \ref{fig:carracing_curves_positve},\ref{fig:carracing_curves_negative},\ref{fig:carracing_curves_composite}. Figure \ref{fig:carracing_curves_positve} demonstrates the previously shown positive intrinsic conditioning results. Figure \ref{fig:carracing_curves_negative} demonstrates the negative intrinsic reward results. Finally, Figure \ref{fig:carracing_curves_composite} provides the composite reward results.

\begin{figure}[!h] 

   \centering
    \begin{subfigure}[t]{0.8\textwidth}
        \centering
        \includegraphics[width=0.9\textwidth, trim={0 0 0 1.2cm}, clip]{Media/Results/Sidewalk/Intrinsic_Only/intrinsic_only_episode_length.png}
        \caption{Episode Length}
        \label{fig:INT_LENGHT}
    \end{subfigure}
~
    \begin{subfigure}[t]{0.8\textwidth}
        \centering
        \includegraphics[width=0.9\textwidth, trim={0 0 0 1.1cm}, clip]{Media/Results/Sidewalk/Intrinsic_Only/intrinsic_only_episodic_reward.png}
        \caption{Episodic Reward}
        \label{fig:INT_EXTRINSIC}
    \end{subfigure}
    ~
    \begin{subfigure}[t]{0.8\textwidth}
        \centering
        \includegraphics[width=0.9\textwidth, trim={0 0 0 1.2cm}, clip]{Media/Results/Sidewalk/Intrinsic_Only/intrinsic_only_intrinsic_reward.png}
        \caption{Episodic Intrinsic}
    \end{subfigure}

    \caption{Training curves for the intrinsic-only condition in the Sidewalk environment across three representative thresholds ($\theta_{\text{thr}} \in \{0.25, 0.60, 0.95\}$), averaged over 5 runs with variance shown as shaded regions. Dashed horizontal lines indicate baseline performance (Base PPO at 116.88, Stimuli at 44.30). (a)~$\theta_{\text{thr}} = 0.60$ maintains stable survival above base PPO, while $\theta_{\text{thr}} = 0.95$ progressively declines below it. (b)~Extrinsic reward remains near zero, confirming the non-descriptive terminal condition. (c)~$\theta_{\text{thr}} = 0.95$ begins with severe intrinsic penalties and only partially recovers, while lower thresholds maintain small penalties near zero.}
    \label{fig:sidewalk_Intrinsic_Only}

\end{figure}

\begin{figure}[!h] 

   \centering
    \begin{subfigure}[t]{0.8\textwidth}
        \centering
        \includegraphics[width=0.9\textwidth, trim={0 0 0 1.2cm}, clip]{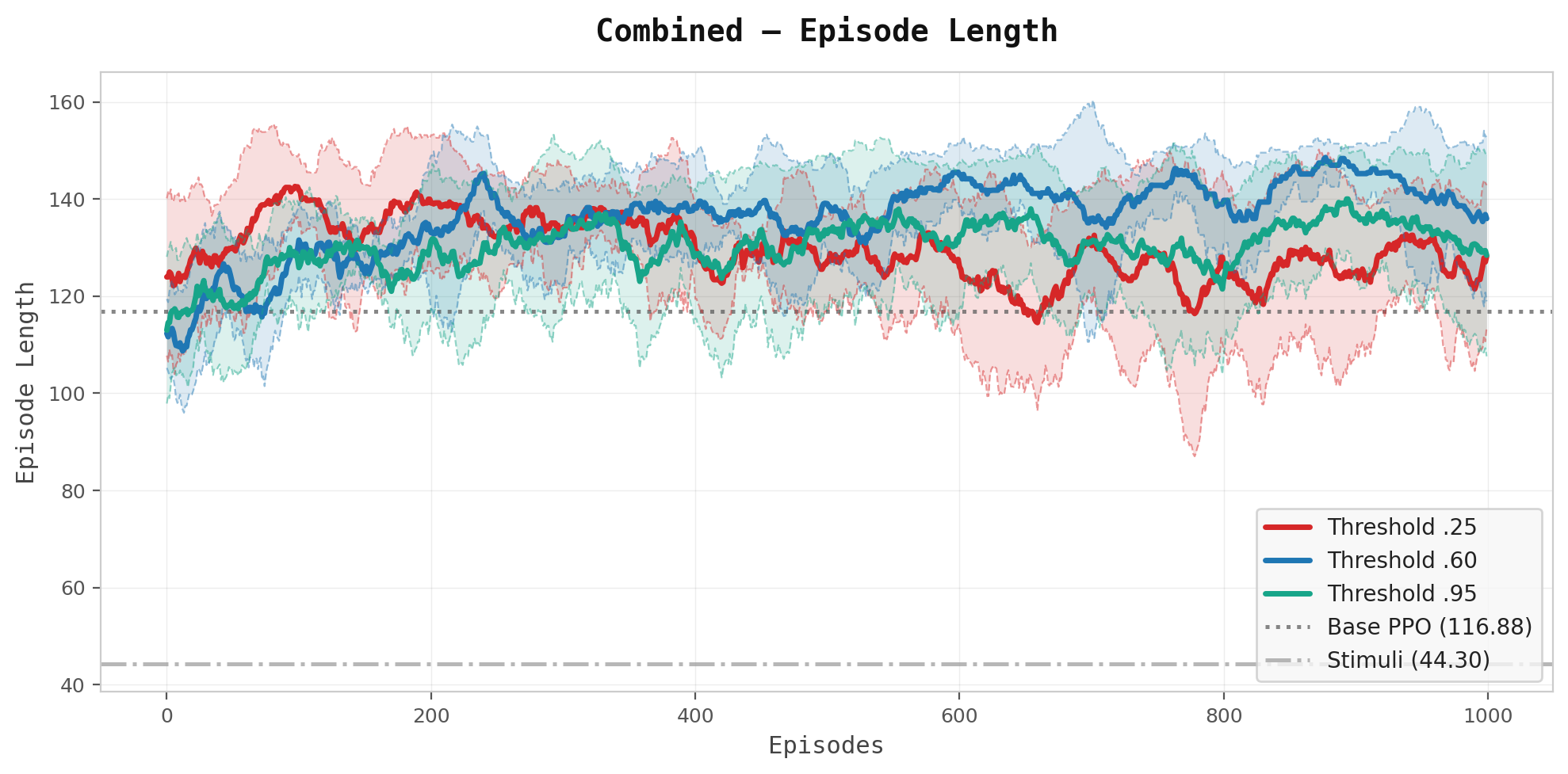}
        \caption{Episode Length}
        \label{fig:INT_LENGHT}
    \end{subfigure}
~
    \begin{subfigure}[t]{0.8\textwidth}
        \centering
        \includegraphics[width=0.9\textwidth, trim={0 0 0 1.1cm}, clip]{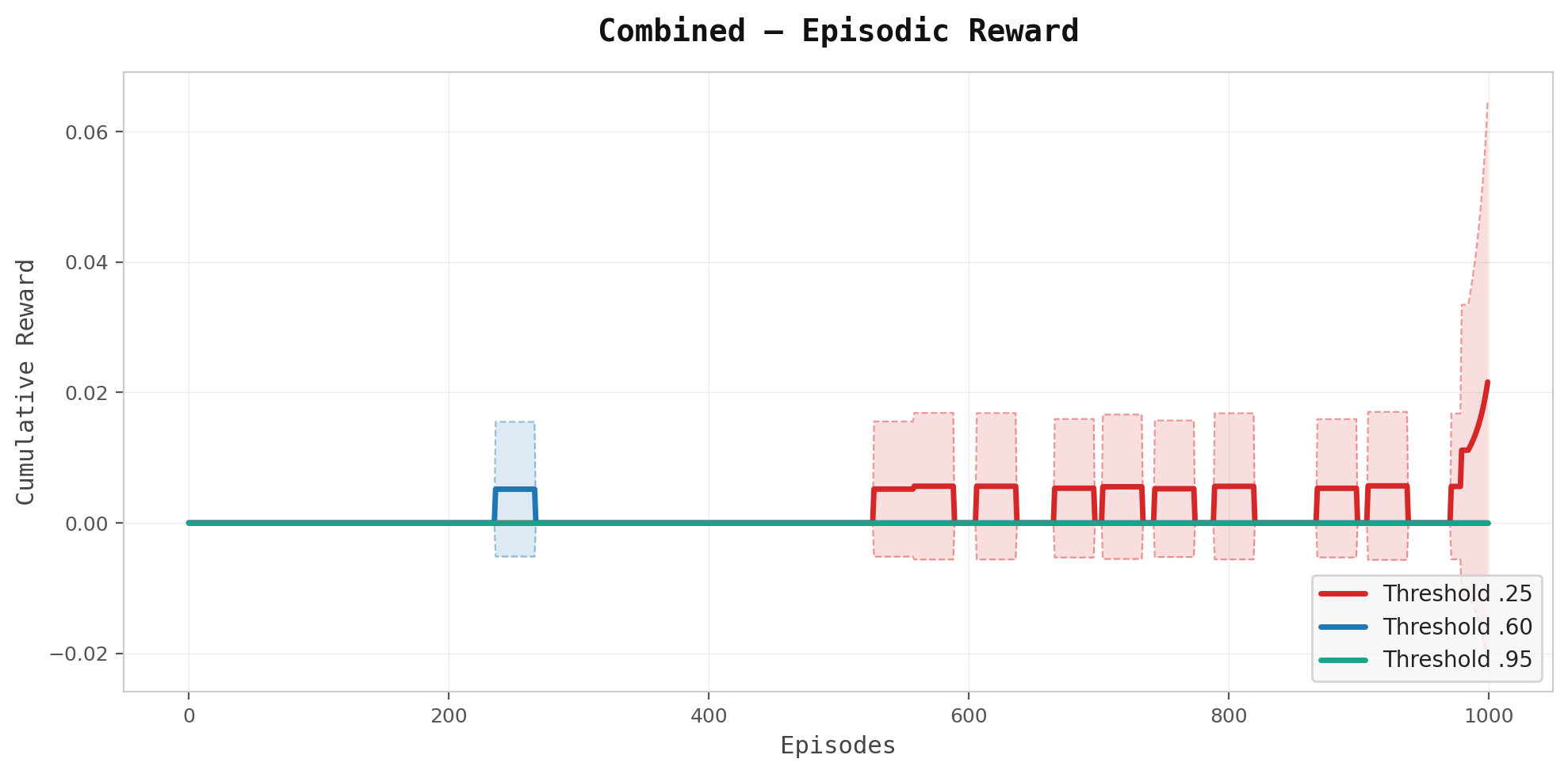}
        \caption{Episodic Reward}
        \label{fig:INT_EXTRINSIC}
    \end{subfigure}
    ~
    \begin{subfigure}[t]{0.8\textwidth}
        \centering
        \includegraphics[width=0.9\textwidth, trim={0 0 0 1.2cm}, clip]{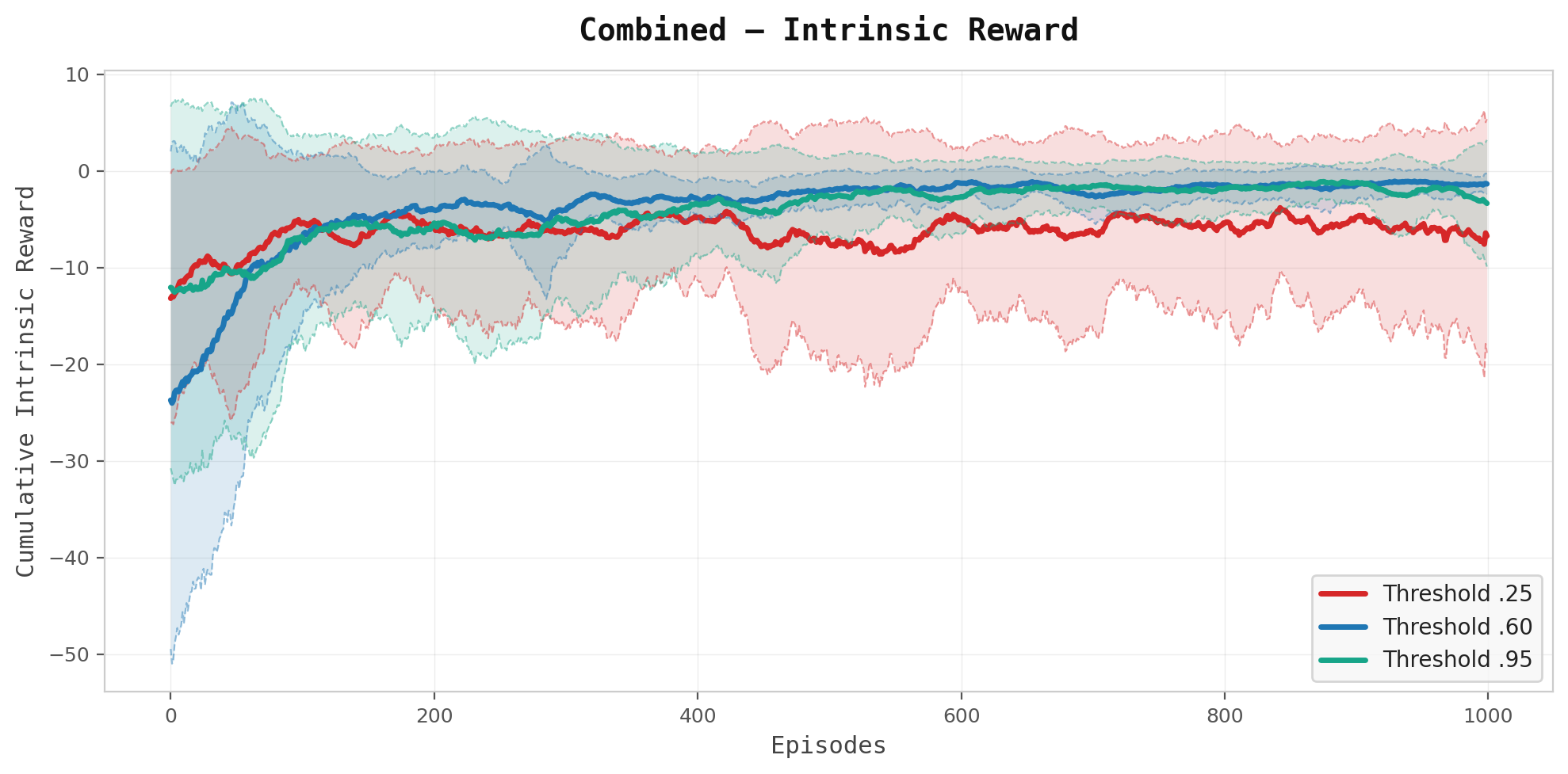}
        \caption{Episodic Intrinsic}
    \end{subfigure}

    \caption{Training curves for the intrinsic and extrinsic condition in the Sidewalk environment across three representative thresholds ($\theta_{\text{thr}} \in \{0.25, 0.60, 0.95\}$), averaged over 5 runs with variance shown as shaded regions. Dashed horizontal lines indicate baseline performance (Base PPO at 116.88, Stimuli at 44.30). (a)~$\theta_{\text{thr}} = 0.60$ maintains stable survival above base PPO, while $\theta_{\text{thr}} = 0.95$ progressively declines below it. (b)~Extrinsic reward remains near zero, confirming the non-descriptive terminal condition except for the .$\theta_{\text{thr}} = 0.25$ which solves it sporadically. (c)~$\theta_{\text{thr}} = 0.95$ begins with severe intrinsic penalties and only partially recovers, while lower thresholds maintain small penalties near zero.}
    \label{fig:sidewalk_Intrinsic_Composite}

\end{figure}

\end{document}